\documentclass{article} 
\usepackage{iclr2027_conference,times}

\usepackage{amsmath,amsfonts,bm}

\def\eqref#1{equation~\ref{#1}}

\def\1{\bm{1}}

\DeclareMathAlphabet{\mathsfit}{\encodingdefault}{\sfdefault}{m}{sl}
\SetMathAlphabet{\mathsfit}{bold}{\encodingdefault}{\sfdefault}{bx}{n}

\usepackage{hyperref}
\usepackage{url}

\usepackage{amsmath,amsfonts}
\usepackage{algorithmic}
\usepackage[ruled,vlined,linesnumbered]{algorithm2e}
\usepackage{array}
\usepackage[caption=false,font=normalsize,labelfont=sf,textfont=sf]{subfig}
\usepackage{textcomp}
\usepackage{stfloats}
\usepackage{url}
\usepackage{verbatim}
\usepackage{graphicx}
\usepackage{cite}

\usepackage{booktabs}       
\usepackage{amsfonts}       
\usepackage{nicefrac}       
\usepackage{microtype}      
\usepackage{xcolor}         
\usepackage{amsmath}
\usepackage{amssymb}
\usepackage{mathtools}
\usepackage{amsthm}

\usepackage{multirow}
\usepackage[table]{xcolor}
\usepackage{color}
\definecolor{myyellow}{RGB}{245, 133, 24} 
\definecolor{myred}{RGB}{178, 121, 162} 
\definecolor{myblue}{RGB}{76, 120, 168} 
\definecolor{mygreen}{RGB}{84, 162, 75} 
\usepackage{graphicx}
\usepackage{adjustbox}
\usepackage{array}
\usepackage{booktabs}
\usepackage{dsfont}
\usepackage{makecell}
\usepackage{wrapfig}
\usepackage{hyperref}
\usepackage{mathrsfs}
\usepackage{xspace}
\usepackage{pifont}

\DeclareRobustCommand{\mycircle}{%
  \raisebox{-0.1\height}{%
    \includegraphics[height=1.6ex]{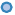}%
  }%
}

\DeclareRobustCommand{\mystar}{%
  \raisebox{-0.35\height}{%
    \includegraphics[height=2.2ex]{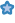}%
  }%
}

\title{Bridging Body and Brain: \\ Gene-Driven Morphology--Control Co-Design}

\iclrfinalcopy

\author{Fu Feng$^{1,2}$,\quad Ruixiao Shi$^{1,2}$,\quad Yucheng Xie$^{1,2}$,\quad Jing Wang$^{1,2*}$,\quad Xin Geng$^{1,2}$\thanks{Co-corresponding author} \\
$^1$ School of Computer Science and Engineering, Southeast University, Nanjing, China \\
$^2$ Key Laboratory of New Generation Artificial Intelligence Technology and Its Interdisciplinary\\
Applications (Southeast University), Ministry of Education, China \\
\texttt{\{fufeng,eric\_xiao,xieyc,wangjing91,xgeng\}@seu.edu.cn} \\
}

\begin{document}

\maketitle
\begin{abstract}
Morphology--control co-design jointly optimizes an agent’s body structure and control policy as an integrated embodied system. 
However, existing methods typically model morphology design and control with separate networks coupled only indirectly through a shared task objective, limiting explicit high-level coordination.
Inspired by natural genes that coordinate biological development, we introduce \textbf{Morphogene}, a compact latent blueprint that bridges an agent's body and brain.
Through AdaConcat, Morphogene jointly conditions morphology and control generation at the limb level, allowing its variations to induce coordinated changes in both components.
Building on this representation, we propose \textbf{GeCode}, which formulates co-design as exploration in the compact Morphogene space.
Each Morphogene anchors a local design region in which nearby body--brain designs are explored, while performance-guided updates move these anchors toward promising regions for more efficient exploration of the broader design space.
This process combines local refinement with global exploration while preserving body--brain compatibility.
Extensive experiments across diverse 2D and 3D co-design tasks demonstrate that GeCode consistently outperforms existing state-of-the-art methods, achieving substantially faster convergence and higher final performance. 
\end{abstract}

\vspace{-0.2in}
\begin{figure*}[ht]
  \centering
  \includegraphics[width=\linewidth]{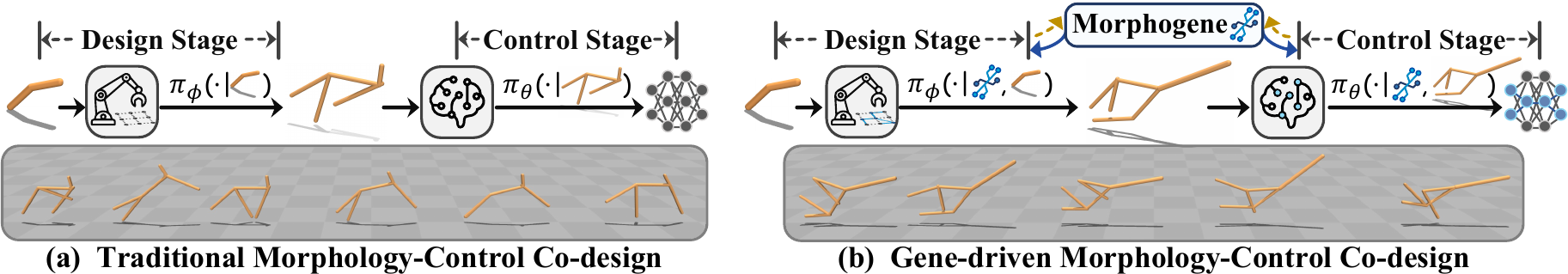}
  \vspace{-0.3in}
  \caption{(a)~Traditional morphology--control co-design employs \textit{separate} morphology-design and control networks coupled primarily through a shared task objective, resulting in weak body--brain coordination.
  (b)~Inspired by biological genes, GeCode introduces Morphogene as a high-level latent blueprint that bridges body and brain.
  By propagating Morphogene variations to both morphology design and controller adaptation,
  GeCode drives their explicit coordination at the representation level while reformulating morphology search as exploration within a continuous latent space.
  }
  \label{fig:main_moti}
  \vspace{-0.1in}
\end{figure*}

\section{Introduction}
Across species in nature, body morphology and neural control develop and evolve together, with the body shaping the physical conditions for behavior and the nervous system adapting to exploit them~\citep{chiel1997brain,bartolozzi2022embodied}. 
Nevertheless, embodied-agent design often treats morphology and control as largely independent components~\citep{xiong2023universal, gupta2021embodied}.
Such one-sided optimization either confines control to the affordances of a fixed body~\citep{haarnoja2024learning,bolotnikova2025optimized} or yields morphologies biased toward a fixed controller~\citep{matthews2023efficient,zhao2020robogrammar}, motivating \textbf{morphology--control co-design} through the joint adaptation of agent's body and brain~\citep{ha2019reinforcement, liu2025embodied,wang2025embodied}.

However, current embodied co-design methods typically employ separate morphology-generation and control-policy networks coupled primarily through the shared task objective (Eq.~(\ref{equ:obj})) \textit{rather than explicit coordination at the representation or decision level}~\citep{yuan2022transformact,lu2025bodygen}. 
Consequently, morphological changes do not directly guide controller adaptation, nor do controller updates explicitly inform morphological design; this weak coupling obscures cross-component credit assignment and leaves body--brain coordination indirect and inadequately modeled~\citep{dai2026stackelberg}.

In nature, genes address an analogous coordination problem by serving as \textcolor{orange}{\textbf{developmental blueprints}} that encode shared high-level programs. Variations in these programs jointly shape morphology, neural circuitry, and behavior beyond adaptation driven solely by environmental feedback~\citep{zador2019critique,fleck2023inferring}.
Inspired by this principle, we introduce \textbf{Morphogene} as a similar \textcolor{orange}{\textbf{developmental blueprint}} for embodied agents.
Rather than representing a single fixed morphology, each Morphogene abstracts high-level structural and geometric characteristics, \textit{anchors} a local region of related body designs, and \textit{jointly conditions} morphology generation and controller adaptation\footnote{The anchoring role of a Morphogene resembles that of a \textbf{\textit{prototype}} in representation learning. Here, however, we use the biologically inspired term \textcolor{orange}{\textbf{\textit{developmental blueprint}}} to emphasize its broader role: beyond serving as a representative embedding, a Morphogene can be adaptively moved through the latent space to guide morphology design while bridging morphology generation and controller adaptation in embodied agents.}.

Specifically, unlike conventional approaches that model body and brain with separate networks~\citep{yuan2022transformact,lu2025bodygen}, a shared Morphogene explicitly couples morphology and control generation at the representation level. Morphogene is injected into both networks through AdaConcat, which transforms the agent-level representation into limb-specific conditions based on each limb's structural and functional role. Consequently, variations in Morphogene induce coherent changes in morphology design and corresponding controller adaptations, thereby bridging body and brain.

Building on Morphogene, we introduce \textbf{GeCode}, a \underline{ge}ne-driven \underline{co}-\underline{de}sign framework that shifts co-design from a discrete morphology-design space~\citep{gupta2021embodied,koike2023simultaneous} to a compact, continuous Morphogene space.
Specifically, GeCode maintains a small set of Morphogenes as design anchors, combining stochastic policy rollouts with a Morphogene-proximity reward to explore high-performing, locally coherent body--brain designs around each anchor.
Performance-guided Morphogene updates subsequently move the anchors toward regions of higher-performing body--brain designs, thereby expanding exploration across the latent space.
This local-to-global strategy enables broad yet targeted exploration without exhaustively evaluating discrete morphologies.

We evaluate GeCode across twelve 2D and 3D design spaces spanning diverse morphologies. GeCode consistently outperforms state-of-the-art methods, achieving on average 2.5$\times$ faster convergence and 69.48\% higher task performance while incurring less than 4.0\% additional wall-clock overhead. 

Our contributions are as follows:
(1) We introduce \textbf{Morphogene}, a high-level latent blueprint that bridges an agent's body and brain, together with AdaConcat for adaptive, limb-aware conditioning of morphology and control.
(2) We propose \textbf{GeCode}, which reformulates morphology--control co-design as gene-driven exploration in the compact Morphogene space, allowing latent updates to induce coordinated body--brain adaptation.
(3) Extensive experiments demonstrate that GeCode achieves state-of-the-art performance with faster convergence and higher final returns.

\vspace{-0.05in}
\section{Related Work}
\vspace{-0.08in}
\paragraph{Morphology--Control Co-design}
Joint morphology--control optimization has emerged as a promising direction in embodied intelligence~\citep{strgar2026accelerated,fay2026house,wang2025embodied,liu2025embodied}. 
Early evolutionary approaches relied on population-based morphology search and repeated controller adaptation, resulting in substantial sampling and computational costs~\citep{Bhatia2021evo, gupta2021embodied,cheney2018scalable}. 
Recent RL-based methods jointly optimize sequential morphology construction and control~\citep{yuan2022transformact,wang2023curriculum,lu2025bodygen}. However, their separate morphology and control networks remain weakly coupled, limiting explicit body--brain coordination.
StackelbergPPO~\citep{dai2026stackelberg} strengthens optimization-level coordination by explicitly modeling morphology--control interdependence, but still leave their representation-level coupling implicit.
In contrast, GeCode makes this coupling explicit through Morphogene, which jointly coordinates morphology and control generation throughout co-design.

\vspace{-0.04in}
Further discussion on \textbf{``Genes'' in Intelligent Agents} and  \textbf{Universal Control} is provided in App.~\ref{app:related}.
\vspace{-0.01in}



\section{Preliminaries}
\label{sec:preli}
\vspace{-0.08in}
\paragraph{Morphology Representation}
An agent's morphology is represented as an undirected graph $\mathcal{M}=(V,E,A)$, where each node $u\in V$ denotes a limb and each edge $e{(u,v)}\in E$ denotes a joint connecting two limbs. The attribute set
$A=\{\Lambda_u^{V}\mid u\in V\}\cup\{\Lambda_e^{E}\mid e\in E\}$
specifies limb properties (e.g., length and size) and joint properties (e.g., rotation ranges and motor torques).

Once the topology and attributes are specified, the simulator (MuJoCo in our implementation) instantiates the morphology. In our design spaces, each limb $u$ is paired with \textit{exactly one} associated joint, denoted by $e{(u,\cdot)}$, allowing their simulator-derived attributes to be combined as
$\mathbf{m}_u=[\mathbf{m}_u^{V}\|\mathbf{m}_{e{(u,\cdot)}}^{E}]\in\mathbb{R}^{d_m}$.
Here, $\mathbf{m}_u^{V}$ encodes fixed limb geometry, such as its reference position and orientation (e.g., \texttt{body\_pos} and \texttt{body\_iquat}), while $\mathbf{m}_{e{(u,\cdot)}}^{E}$ encodes the corresponding joint geometry (e.g., \texttt{jnt\_pos}). The complete vector representation is
$\mathcal{M}^{\mathrm{vec}}=[\mathbf{m}_u]_{u\in V}$.
Thus, $\mathcal{M}$ and $\mathcal{M}^{\mathrm{vec}}$ describe the same morphology at the graph and vector levels, respectively, with the former used for sequential morphology design and the latter for constructing the Morphogene space.

\paragraph{Morphology-Control Co-Design Optimization}
Morphology--control co-design optimizes an agent's body structure and control policy through a \textbf{\textit{Design Stage}} $\Phi_{\mathrm{des}}$ and a \textbf{\textit{Control Stage}} $\Phi_{\mathrm{ctrl}}$. 
During morphology design $\Phi_{\mathrm{des}}$, the morphology graph $\mathcal{M}_t=(V_t,E_t,A_t)$ is sequentially updated over $T_{\mathrm{topo}}$ topology design steps followed by $T_{\mathrm{attr}}$ attribute design steps.
Each update follows
\begin{equation*}
a_t^{\mathrm{des}}\sim\pi_{\phi}^{\mathrm{des}}(\cdot\mid\mathcal{M}_t),\qquad \mathcal{M}_{t+1}\sim P^{\mathrm{des}}(\cdot\mid\mathcal{M}_t,a_t^{\mathrm{des}}),\quad 0 < t \leq T_{\mathrm{des}},
\end{equation*}
where $T_{\mathrm{des}}=T_{\mathrm{topo}}+T_{\mathrm{attr}}$.
$\pi_{\phi}^{\mathrm{des}}$ is the morphology-design policy, and $P^{\mathrm{des}}$ is the corresponding transition function. 
The morphology-editing action $a_t^{\mathrm{des}}$ specifies structural operations (e.g., adding or removing limbs) during topology design and physical-attribute updates (e.g., adjusting limb length) during attribute design.
The resulting action sequence yields the final morphology $\mathcal{M}^{\star}=\mathcal{M}_{T_{\mathrm{des}}}$.

During the control stage $\Phi_{\mathrm{ctrl}}$, the morphology $\mathcal{M}^{\star}$ remains fixed while the controller interacts with the environment to maximize the expected task return. At each control step, the interaction follows
\begin{equation*}
a_t^{\mathrm{ctrl}}\sim\pi_\theta^{\mathrm{ctrl}}(\cdot\mid s_t^{\mathrm{ctrl}};\mathcal{M}^{\star}),\qquad s_{t+1}^{\mathrm{ctrl}}\sim P^{\mathrm{ctrl}}(\cdot\mid s_t^{\mathrm{ctrl}},a_t^{\mathrm{ctrl}};\mathcal{M}^{\star}),\qquad t > T_{\mathrm{des}}.
\end{equation*}
Here, $s_t^{\mathrm{ctrl}}$ denotes the environment observation received by the controller, while $\pi_\theta^{\mathrm{ctrl}}$ and $P^{\mathrm{ctrl}}$ denote the control policy and environment transition function under $\mathcal{M}^{\star}$, respectively.

The morphology--control co-design objective jointly optimizes the design and control policy parameters to maximize the expected discounted return:
\begin{equation}
\max_{\phi,\theta}\; J(\phi,\theta)
=
\mathbb{E}\left[\sum_{t=T_{\mathrm{des}}+1}^{\infty}\gamma^{t-T_{\mathrm{des}}-1}r_t^{\mathrm{env}}\right].
\label{equ:obj}
\end{equation}
Here, $\gamma$ is the discount factor, and $r_t^{\mathrm{env}}$ denotes the task reward provided by the environment (App.~\ref{app:env}).

Existing methods~\citep{lu2025bodygen,yuan2022transformact} use separate morphology and control policies coupled mainly through the shared objective in Eq.~(\ref{equ:obj}), leaving no explicit mechanism to coordinate morphological changes with controller adaptation.
To explicitly couple these two components, we formulate morphology generation and controller adaptation as Morphogene-conditioned processes (Sec.~\ref{sec:ada}) and jointly optimize them under the augmented objective in Eq.~(\ref{equ:joint_objective}).

\paragraph{Transformer-based Morphology Design and Control}
An agent is represented by an ordered sequence of $N_L$ limb-wise states, where $\mathbf{s}_i\in\mathbb{R}^{d_s}$ is the proprioceptive observation of limb $l_i$. Each state is projected into an embedding space, and the resulting limb tokens are assembled as
\begin{equation}
\mathbf{h}_{i}
=
f_{\mathrm{emb}}(\mathbf{s}_{i}),
\quad i=1,\ldots,N_L,
\qquad
\mathbf{H}^{(0)}
=
\left[
\mathbf{h}_1;
\cdots;
\mathbf{h}_{N_L}
\right]
\label{equ:transformer}
\end{equation}
where $f_{\mathrm{emb}}(\cdot)$ is the shared embedding function~\citep{guptametamorph}.

The Transformer encoder comprises $L$ pre-normalized blocks, each applying layer normalization~(LN)~\citep{xiong2020layer} before multi-head self-attention (MSA)~\citep{vaswani2017attention} and a feed-forward network (FFN), followed by residual updates:
\begin{equation}
\widetilde{\mathbf{H}}^{(\ell)}
=
\mathbf{H}^{(\ell-1)}
+
\operatorname{MSA}_{\ell}
\!\left(
\operatorname{LN}_{\ell,1}
\!\left(\mathbf{H}^{(\ell-1)}\right)
\right),
\;
\mathbf{H}^{(\ell)}
=
\widetilde{\mathbf{H}}^{(\ell)}
+
\operatorname{FFN}_{\ell}
\!\left(
\operatorname{LN}_{\ell,2}
\!\left(\widetilde{\mathbf{H}}^{(\ell)}\right)
\right),
\;
\ell=1,\ldots,L.
\label{equ:msa}
\end{equation}
The final token representation $\mathbf{H}^{(L)}$ is mapped by a linear action head to the distribution parameters $\mathbf{z}_{\mathrm{act}}=f_{\mathrm{act}}(\mathbf{H}^{(L)})$. During the design stage $\Phi_{\mathrm{des}}$, discrete actions follow a categorical distribution parameterized by $\operatorname{softmax}(\mathbf{z}_{\mathrm{act}})$. During the control stage $\Phi_{\mathrm{ctrl}}$, continuous actions follow a Gaussian distribution $\mathcal{N}(\boldsymbol{\mu}_{\mathrm{act}},\boldsymbol{\Sigma}_{\mathrm{act}})$, whose mean and covariance are parameterized by $\mathbf{z}_{\mathrm{act}}$.

\section{Method}
\subsection{Morphogene: A Latent Blueprint for Body--Brain Coordination}
\label{sec:morphgene}
\paragraph{Morphology Encoding and Morphogene Representation}
We define Morphogene as a compact latent blueprint that guides morphology generation and compatible control.
To ground this blueprint in the underlying body structure, we learn a latent space that captures the correlated structural and geometric factors of morphology.
As illustrated in Figure~\ref{fig:RAE}(a), the morphology encoder $E_{\psi}$ maps an instantiated morphology $\mathcal{M}^{\mathrm{vec}}$ to its morphology code $\mathbf{z}^{\mathrm{morph}}$:
\begin{equation}
\mathbf{z}^{\mathrm{morph}}
=
E_{\psi}\!\left(\mathcal{M}^{\mathrm{vec}}\right),
\qquad
\mathbf{z}^{\mathrm{morph}}\in\mathbb{R}^{d_g},
\label{eq:morphogene-encoder}
\end{equation}
Unlike \textcolor{blue}{\textbf{morphology code $\mathbf{z}^{\mathrm{morph}}$}}, which encodes \textcolor{blue}{\textbf{a specific realized morphology}}, 
a \textcolor{orange}{\textbf{Morphogene $\mathcal{G}\in\mathbb{R}^{d_g}$}} serves as \textcolor{orange}{\textbf{a searchable anchor}} in the same latent space. GeCode then iteratively updates these anchors to drive exploration of coordinated body--brain designs (Sec.~\ref{sec:gecode}).

\paragraph{RAE-based Morphogene Space Construction}
The Morphogene space is learned using a regularized autoencoder (RAE), whose decoder $D_{\psi}$ reconstructs morphological attributes and predicts limb presence, thereby preserving structural and geometric information:
\begin{equation}
(\widehat{\mathcal{M}}^{\mathrm{vec}},\widehat{\boldsymbol{\chi}})
=
D_{\psi}(\mathbf{z}^{\mathrm{morph}})
=
D_{\psi}\!\left(E_{\psi}(\mathcal{M}^{\mathrm{vec}})\right),
\label{eq:morphogene-decoder}
\end{equation}
Here, $\widehat{\boldsymbol{\chi}}$ denotes the predicted limb-presence mask, which determines the number of limbs in the reconstructed morphology. 
The RAE is trained using a masked, regularized reconstruction objective:
\begin{equation}
\mathcal{L}_{\mathrm{rec}}
=
\mathcal{L}_{\mathrm{attr}}
+
\lambda_{\mathrm{mask}}\mathcal{L}_{\mathrm{mask}}
+
\lambda_{\mathrm{reg}}\mathcal{L}_{\mathrm{reg}},
\label{eq:rae-objective}
\end{equation}
where $\mathcal{L}_{\mathrm{attr}}
=
\frac{\|\boldsymbol{\chi}\odot(\widehat{\mathcal{M}}^{\mathrm{vec}}-\mathcal{M}^{\mathrm{vec}})\|_2^2}
{\|\boldsymbol{\chi}\|_1}$
measures reconstruction error over valid morphology components,
$\mathcal{L}_{\mathrm{mask}}
=
\operatorname{BCE}(\widehat{\boldsymbol{\chi}},\boldsymbol{\chi})$
supervises limb-presence prediction, and
$\mathcal{L}_{\mathrm{reg}}
=
\|\mathbf{z}^{\mathrm{morph}}\|_2^2$
regularizes the morphology codes to promote a compact Morphogene space.

\paragraph{Structural Constraints on the Morphogene Space}
While the reconstruction objective preserves the information required for morphology recovery, it does not explicitly impose meaningful geometric structure on the Morphogene space.
We therefore regularize RAE with a topology-aware contrastive loss  $\mathcal{L}_{\mathrm{topo}}$~(Eq.~(\ref{eq:topology_contrastive})) and a tree-structured metric loss $\mathcal{L}_{\mathrm{tree}}$~(Eq.~(\ref{eq:tree_metric})), detailed in App.~\ref{app:morphogene_structural_losses}. 

Specifically, $\mathcal{L}_{\mathrm{topo}}$ clusters morphology codes with identical topologies while separating those with different topologies, whereas $\mathcal{L}_{\mathrm{tree}}$ enforces relative-distance margins between structurally similar and dissimilar morphologies. 
Together, these losses align latent geometry with morphological relationships, thereby enabling smooth Morphogene exploration, as visualized in Figure~\ref{fig:latent_space}(b).

\subsection{Morphogene-Conditioned Morphology Design and Control}
\label{sec:ada}
GeCode jointly conditions the morphology-design policy $\pi_{\phi}^{\mathrm{des}}(\cdot\mid\mathcal{G})$ and control policy $\pi_{\theta}^{\mathrm{ctrl}}(\cdot\mid\mathcal{G})$ on a shared Morphogene $\mathcal{G}$, initially sampled at random from the learned Morphogene space, thereby explicitly coupling body and brain.
Following Eq.~(\ref{equ:transformer}), the state of limb $l_i$ is first mapped to a stage-specific feature $\mathbf{h}_i=f_{\mathrm{emb}}(\mathbf{s}_{i};\Phi_\star)$, where $\Phi_\star\in\{\Phi_{\mathrm{des}},\Phi_{\mathrm{ctrl}}\}$ denotes the active stage.

\begin{figure*}[tb]
  \centering
  \includegraphics[width=\linewidth]{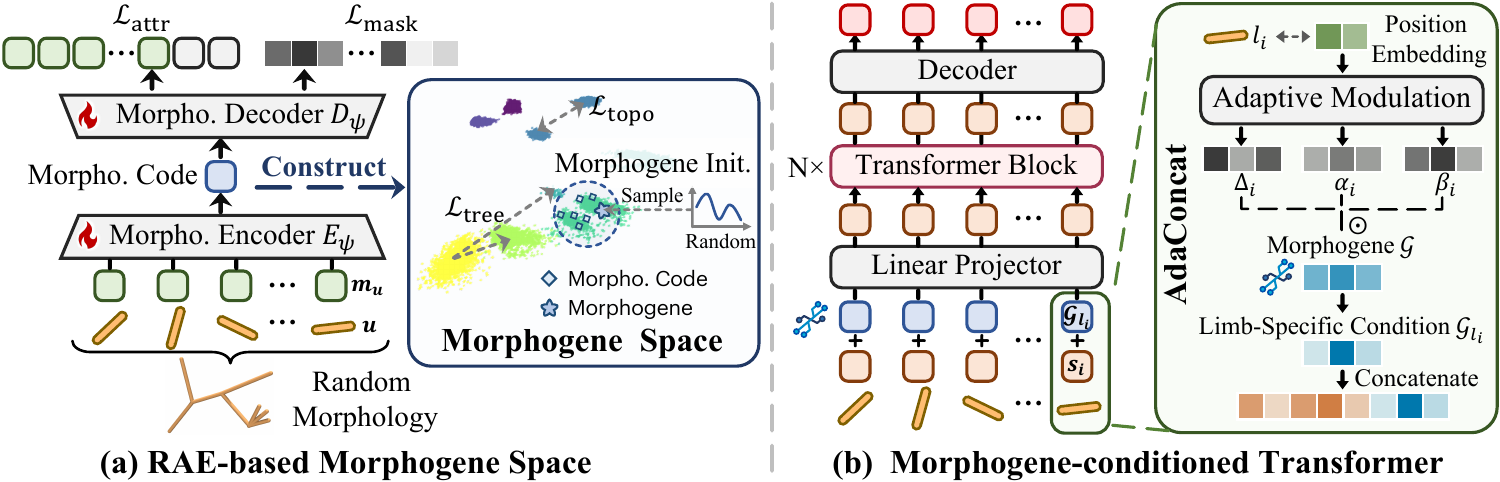}
  \vspace{-0.23in}
  \caption{\textbf{Overview of Morphogene space construction and conditioning.}
  (a)~An RAE encodes randomly generated morphologies into morphology codes, forming a continuous latent space in which Morphogenes serve as searchable anchors.
  (b)~AdaConcat adapts an agent-level Morphogene into limb-specific conditions for the morphology-design and control networks.}
  \label{fig:RAE}
  \vspace{-0.05in}
\end{figure*}

\paragraph{AdaConcat for Limb-wise Condition Injection} 
As the agent-level Morphogene captures the morphology as a whole, uniformly injecting it across limbs cannot reflect their distinct structural and functional roles. 
We therefore introduce AdaConcat to adapt $\mathcal{G}$ for limb-specific conditioning:
\begin{equation}
\mathbf{e}_i
=
\operatorname{Emb}_{\mathrm{pos}}(i)
\in\mathbb{R}^{d_{\mathrm{pos}}},
\qquad
[\boldsymbol{\Delta}_i,\boldsymbol{\alpha}_i,\boldsymbol{\beta}_i]
=
f_{\mathrm{Ada}}(\mathbf{e}_i;\Phi_{\star})
\in\mathbb{R}^{3d_g},
\end{equation}
where $\mathbf{e}_i$ is the positional embedding of limb $l_i$, and $f_{\mathrm{Ada}}:\mathbb{R}^{d_\mathrm{pos}}\rightarrow\mathbb{R}^{3d_g}$ is the modulation network.
For limb $l_i$, AdaConcat transforms $\mathcal{G}$ into the corresponding limb-specific condition $\mathcal{G}_{l_i}$ as
\begin{equation}
\mathcal{G}_{l_i}
=
\boldsymbol{\Delta}_i\odot
\left[
(\mathbf{1}+\boldsymbol{\alpha}_i)\odot\mathcal{G}
+
\boldsymbol{\beta}_i
\right],
\label{equ:adaconcat}
\end{equation}
where $\odot$ denotes element-wise multiplication. See Figure~\ref{fig:RAE}(b) for an illustration.

The limb-specific condition $\mathcal{G}_{l_i}$ is concatenated with its corresponding limb feature to form the conditioned token $\mathbf{h}'_i=[\mathbf{h}_i \,\|\, \mathcal{G}_{l_i}]$. 
The conditioned tokens are assembled as 
$\mathbf{H}'^{(0)}=[\mathbf{h}_1';\ldots;\mathbf{h}_{N_L}']$ and processed by the Transformer blocks in Eq.~(\ref{equ:msa}), yielding a morphology-editing action $a^{\mathrm{des}}$ during $\Phi_{\mathrm{des}}$ or a control action $a^{\mathrm{ctrl}}$ during $\Phi_{\mathrm{ctrl}}$:
\begin{equation}
a^{\mathrm{des}}
\sim
\pi_{\phi}^{\mathrm{des}}
(\cdot\mid\mathcal{M};\mathcal{G}),\ \Phi_{\star}=\Phi_{\mathrm{des}},
\qquad
a^{\mathrm{ctrl}}
\sim
\pi_{\theta}^{\mathrm{ctrl}}
(\cdot\mid s^{\mathrm{ctrl}};\mathcal{M}^{\star},\mathcal{G}),\ \Phi_{\star}=\Phi_{\mathrm{ctrl}}.
\label{equ:gene_action}
\end{equation}
In this way, the shared Morphogene aligns morphology generation with controller adaptation, allowing its variations to induce coherent body--brain changes for explicit morphology--control co-design.

\vspace{-0.1in}
\paragraph{Morphogene-Guided Local Exploration}
During co-design, each Morphogene anchors a local design region for exploring nearby high-performing morphology variants. GeCode promotes such exploration by augmenting the task objective in Eq.~(\ref{equ:obj}) with a Morphogene-proximity reward:
\begin{equation}
r_t^{\mathrm{prox}}
=
\lambda_{\mathrm{hit}}\,
\mathbb{I}[t=T_{\mathrm{des}}]\,
\mathbb{I}[\operatorname{sim}(\mathbf{z}_t^{\mathrm{morph}},\mathcal{G})\geq\tau_{\mathrm{sim}}],
\qquad
\mathbb{I}[\mathrm{True}]=1,\;\; 
\mathbb{I}[\mathrm{False}]=0.
\label{equ:proximity_reward}
\end{equation}
At the end of $\Phi_{\mathrm{des}}$, the generated morphology $\mathcal{M}^{\star}$ is mapped to its morphology code $\mathbf{z}^{\mathrm{morph}}_{T_{\mathrm{des}}}$ using Eq.~(\ref{eq:morphogene-encoder}). 
A reward of $\lambda_{\mathrm{hit}}$ is assigned when the cosine similarity $\operatorname{sim}(\cdot, \cdot)$ between the morphology code and its corresponding Morphogene $\mathcal{G}$ exceeds $\tau_{\mathrm{sim}}$.
Thus, the joint training objective is
\begin{equation}
\max_{\phi,\theta}\; J(\phi,\theta)
=
\mathbb{E}
\left[
\sum_{t=1}^{T_{\mathrm{des}}}
r_t^{\mathrm{prox}}
+
\sum_{t=T_{\mathrm{des}}+1}^{\infty}
\gamma^{\,t-T_{\mathrm{des}}-1}\,r_t^{\mathrm{env}}
\right].
\label{equ:joint_objective}
\end{equation}
Consequently, the morphology-design policy $\pi_{\phi}^{\mathrm{des}}$ is optimized to maximize task return within the $\tau_{\mathrm{sim}}$-neighborhood of its conditioning Morphogene.
This local search yields high-performing candidates for Morphogene updates, which subsequently drive global exploration of the latent space.

\subsection{GeCode: Morphogene-Driven Co-design Optimization}
\label{sec:gecode}
\begin{figure*}[tb]
  \centering
  \includegraphics[width=\linewidth]{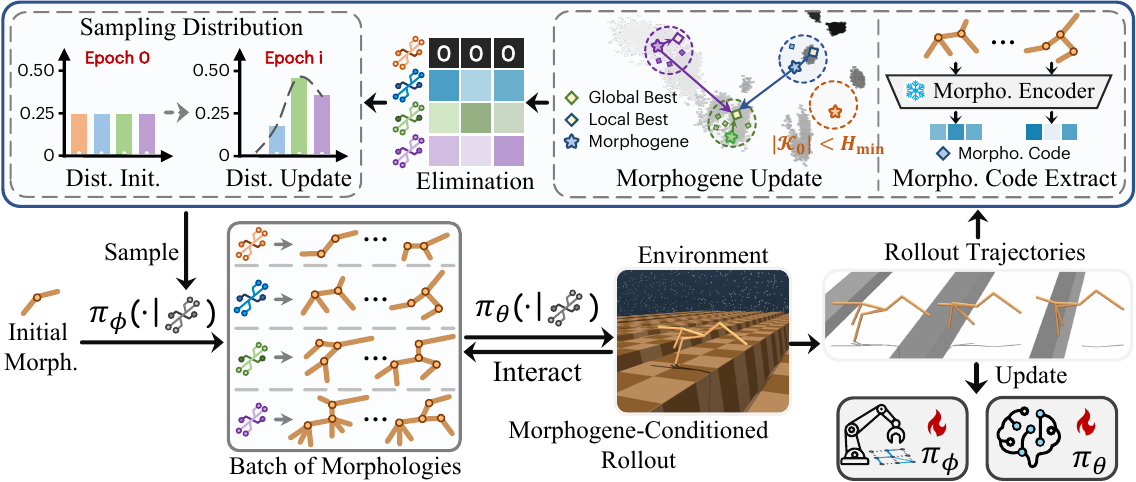}
  \vspace{-0.3in}
  \caption{\textbf{Overview of GeCode.} 
  GeCode reformulates morphology--control co-design as exploration over the compact Morphogene space, using a small set of Morphogenes as searchable design anchors.
  At each co-design epoch, sampled anchors jointly condition morphology generation and controller adaptation, while stochastic rollouts guided by a proximity objective explore their local neighborhoods. Rollout performance subsequently drives Morphogene updates and adaptive resampling, propagating high-performing design information across the set and enabling global latent-space exploration.}
  \label{fig:main_method}
  \vspace{-0.1in}
\end{figure*}

GeCode performs morphology--control co-design in the compact Morphogene space, where each Morphogene serves as a design anchor for body--brain coordination (Figure~\ref{fig:main_method}; Algorithm~\ref{alg:gecode}).
These anchors are updated using existing rollout statistics to efficiently explore diverse body--brain designs, \textit{requiring neither additional environment interactions nor gradient-based latent optimization}.

\paragraph{Morphogene Initialization}
GeCode initializes a small set of Morphogenes, $\mathcal{P}_{\mathcal{G}}=\{\mathcal{G}_i\}_{i=1}^{N_G}$, by randomly and independently sampling from the learned RAE latent distribution. 
These Morphogenes are then iteratively refined throughout co-design, \textit{with no new Morphogenes introduced thereafter}.

\paragraph{Morphogene-Conditioned Rollout}
At each co-design epoch, each rollout first samples a Morphogene $\mathcal{G}_i\in\mathcal{P}_{\mathcal{G}}$ according to $i\sim\mathbf{p}$, where $\mathbf{p}=(p_1,\ldots,p_{N_G})$ is the \textbf{performance-adaptive sampling distribution}. It is initialized only once at epoch $0$ with $p_i=\frac{1}{N_G}$ and iteratively updated after every subsequent epoch according to Eq.~(\ref{equ:update_P}).
The sampled $\mathcal{G}_i$ remains fixed throughout the rollout and conditions both the morphology-design and control policies through AdaConcat (Figure~\ref{fig:RAE}(b)).

Because the morphology-design actions $a^{\mathrm{des}}$ in Eq.~(\ref{equ:gene_action}) are sampled stochastically, $K$ rollouts conditioned on the same $\mathcal{G}_i$ may produce distinct morphology variants $\{\mathcal{M}_{i,k}\}_{k=1}^{K}$ within the local design region anchored by $\mathcal{G}_i$, thereby enabling \textbf{\textit{local exploration}} around the anchor.

\paragraph{Morphogene Update}
After each epoch's rollouts, GeCode updates each Morphogene using the morphology codes of the best-performing design \textit{within its local region} and \textit{across the entire set}, referred to as the local and global best, respectively.
The former preserves anchor-specific morphological characteristics, whereas the latter propagates the best-performing body--brain design across Morphogenes, thereby driving \textbf{\textit{global exploration}} of the latent design space (see Figure~\ref{fig:latent_space}(b)). 

Specifically, each morphology $\mathcal{M}_{i,k}$ generated by rollout $k$ conditioned on Morphogene $\mathcal{G}_i$ is mapped to its morphology code
$\mathbf{z}_{i,k}^{\mathrm{morph}}$ according to Eq.~(\ref{eq:morphogene-encoder}). To preserve locality around $\mathcal{G}_i$, we retain the rollout indices $\mathcal{K}_i=\{k\mid\operatorname{sim}(\mathbf{z}_{i,k}^{\mathrm{morph}}, \mathcal{G}_i)\geq\tau_{\mathrm{sim}}\}$, where $\tau_{\mathrm{sim}}$ filters out morphology codes that deviate excessively from their conditioning Morphogenes. The retained rollouts across all Morphogenes are collected in $\mathcal{K}=\{(i,k)\mid k\in\mathcal{K}_i,\ i=1,\ldots,N_G\}$.

Let $\mathbf{z}_{i}^{\mathrm{loc}}$ and $\mathbf{z}^{\mathrm{glob}}$ denote the codes of the morphologies produced by the highest-return rollouts in $\mathcal{K}_i$ and $\mathcal{K}$, with corresponding returns $R_i^{\mathrm{loc}}$ and $R^{\mathrm{glob}}$, respectively. Their weighted combination defines the update target for $\mathcal{G}_i$:
\begin{equation}
\overline{\mathcal{G}}_i = 
(1-\beta_i)\mathbf{z}_{i}^{\mathrm{loc}}
+
\beta_i\mathbf{z}^{\mathrm{glob}},
\qquad
\mathcal{G}_i
\leftarrow
(1-\eta)\mathcal{G}_i
+
\eta\overline{\mathcal{G}}_i.
\label{equ:update}
\end{equation}
Here, $\beta_i=
\min(
1,\,
\tfrac{R^{\mathrm{glob}}-R_i^{\mathrm{loc}}}
{\max(\lvert R^{\mathrm{glob}}\rvert,1)}
)$
adaptively balances global guidance across Morphogenes, while $\eta$ controls the Morphogene update rate.



\paragraph{Morphogene Elimination}
At the end of each epoch, a Morphogene $\mathcal{G}_i$ is retained only if at least $H_{\min}$ of its conditioned rollouts generate morphologies within its local region.
Accordingly, the active Morphogene set is updated as
\begin{equation}
\mathcal{P}_{\mathcal{G}}
=
\left\{
\mathcal{G}_i
\;\middle|\;
\mathcal{G}_i\in\mathcal{P}_{\mathcal{G}},
\;
|\mathcal{K}_i|\geq H_{\min}
\right\},
\label{equ:eliminate}
\end{equation}
This procedure progressively eliminates Morphogenes that cannot be reliably realized by the morphology-design policy until only one remains.


\paragraph{Adaptive Morphogene Sampling}
After eliminating Morphogenes, GeCode sets their next-epoch sampling probabilities $\mathbf{p}$ using the mean return of the top-$N_{\mathrm{top}}$ rollouts for each retained $\mathcal{G}_i$:
\begin{equation}
q_i
=
\operatorname{AvgTop}
\left(
\{R_{i,k}\}_{k\in\mathcal{K}_i},
N_{\mathrm{top}}
\right),
\qquad
\mathbf{p}
=
\operatorname{softmax}
\left(
\mathbf{q}
\right).
\label{equ:update_P}
\end{equation}
Here, $R_{i,k}$ denotes the return of rollout $k$ conditioned on $\mathcal{G}_i$, with $N_{\mathrm{top}}\leq H_{\min}$.

Overall, GeCode coordinates global exploration through Morphogene updates with local exploration through stochastic Morphogene-conditioned rollouts. By operating on low-dimensional Morphogenes and reusing policy-training rollouts, it requires only lightweight vector operations, enabling efficient Morphogene-driven morphology--control co-design.

\section{Experiments}
Our experiments evaluate whether Morphogene effectively bridges an agent's body and brain and whether GeCode leverages this coordination to achieve sample-efficient, high-performing morphology--control co-design. We further analyze the mechanisms underlying these gains.

\paragraph{Environments}
Following prior work~\citep{lu2025bodygen,dai2026stackelberg}, we evaluate GeCode across twelve MuJoCo environments~\citep{todorov2012mujoco}: \textsc{Crawler}, \textsc{Stepper}, \textsc{Pusher}, \textsc{TerrainCrosser}, \textsc{Cheetah}, \textsc{Swimmer}, \textsc{Glider-(Regular, Medium, Hard)} and \textsc{Walker-(Regular, Medium, Hard)}.
These 2D and 3D benchmarks cover locomotion and manipulation across diverse design spaces and terrains. Further details are provided in App.~\ref{app:env}.

\subsection{Comparison with Baselines}
\begin{figure*}[tb]
  \centering
  \includegraphics[width=\linewidth]{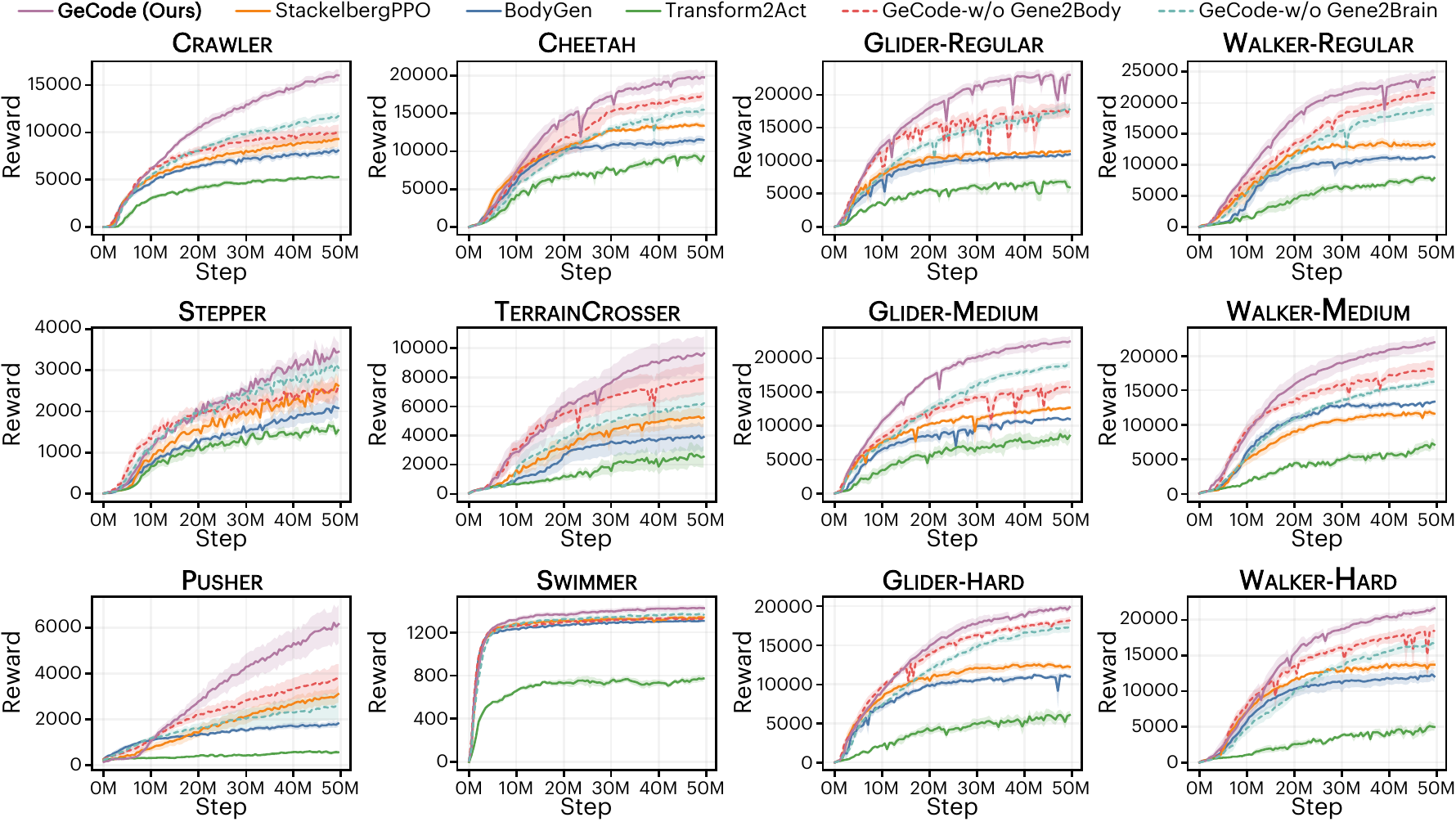}
  \vspace{-0.2in}
  \caption{Performance comparison of GeCode with state-of-the-art baselines, including StackelbergPPO~\citep{dai2026stackelberg}, BodyGen~\citep{lu2025bodygen}, and Transform2Act~\citep{yuan2022transformact}, and its ablations, GeCode-w/o Gene2Body and GeCode-w/o Gene2Brain, across twelve co-design environments. Shaded regions indicate the standard error over five random seeds.}
  \label{fig:main_result}
\end{figure*}

\paragraph{Baselines} 
We build GeCode on \textbf{(1) BodyGen}~\citep{lu2025bodygen}, a PPO-based Transformer framework for morphology--control co-design. GeCode augments this backbone with Morphogenes that bridge body generation and controller adaptation, while Morphogene updates drive efficient exploration of coordinated body--brain designs. BodyGen therefore serves as a direct baseline.
We further compare GeCode with \textbf{(2) StackelbergPPO}~\citep{dai2026stackelberg}, which formulates co-design as a Stackelberg game to explicitly model the optimization interdependence between morphology and control, and \textbf{(3)~Transform2Act}~\citep{yuan2022transformact}, which trains separate GNN-based policies within a unified PPO framework. 
Additional comparisons with evolutionary approaches such as \textbf{(4) NGE}~\citep{wang2018neural} are reported in Table~\ref{tab:main_results} of App.~\ref{app:qua_result}. Implementation details are provided in App.~\ref{app:imp_detail}.

As shown in Figure~\ref{fig:main_result}, GeCode consistently outperforms all baselines by substantial margins, achieving an average final return approximately 69.48\% higher than that of StackelbergPPO, a state-of-the-art co-design method.
These gains extend beyond relatively simple tasks, such as \textsc{Glider-Regular} and \textsc{Walker-Regular}, and become more pronounced in tasks with larger morphology spaces, including \textsc{Crawler}, \textsc{Glider-Hard}, and \textsc{Walker-Hard}.
This trend highlights how GeCode leverages Morphogene-guided exploration in a compact latent space to reduce effective search complexity as the morphology space expands.
Performance gains on complex-terrain tasks such as \textsc{Stepper} and \textsc{TerrainCrosser} further highlight Morphogene's role as a body--brain bridge, integrating proprioception with morphology-dependent contact dynamics to improve control adaptation.
Representative morphologies designed by GeCode are shown in Figure~\ref{fig:update}(a).


\subsection{Ablation and Analysis}
In this section, we investigate five questions:
(1)~how Morphogene \textbf{\textit{contributes separately}} to morphology design and control; 
(2)~how its \textit{\textbf{conditioning mechanism}} contributes to these benefits; 
(3)~how Morphogene \textit{\textbf{update}} and \textit{\textbf{set size}} affect co-design; 
(4)~whether \textit{\textbf{structural regularization}} of the RAE latent space is necessary; 
and (5)~what \textit{\textbf{computational overhead}} the resulting procedure incurs.

\vspace{-0.05in}
\paragraph{Ablation of Morphogene Conditioning on Body and Brain}
Morphogene bridges body and brain by jointly conditioning morphology generation and controller adaptation. 
To isolate its role in each component, we remove Morphogene conditioning from either the morphology-design network (w/o Gene2Body) or the control network (w/o Gene2Brain). Figure~\ref{fig:main_result} shows that both ablations consistently degrade performance, confirming their complementary contributions.
Specifically, conditioning morphology generation on Morphogene is particularly important in morphology-intensive tasks with large, coupled design spaces, such as \textsc{Crawler} and \textsc{Stepper}, where discovering a viable body is prerequisite to effective control. In simpler tasks with lower morphological complexity, such as \textsc{Cheetah} and \textsc{Walker-Regular}, conditioning the controller becomes more influential by enabling morphology-aware action adaptation.

\begin{table*}[t]
\centering
\setlength{\tabcolsep}{1. mm}
\caption{Performance comparison of different Morphogene injection mechanisms. Results show mean episode rewards and standard errors over five random seeds.}
\vspace{0.02in}
\label{tab:adaconcat}
\resizebox{\linewidth}{!}{
\begin{tabular}{@{}lcccccc@{}}
\toprule
& \textbf{\textsc{Crawler}} & \textbf{\textsc{Stepper}} & \textbf{\textsc{Pusher}} & \textbf{\textsc{Cheetah}} & \textbf{\textsc{TerrainCrosser}} & \textbf{\textsc{Swimmer}}  \\
\midrule

w/ Direct Concat
& 14049.3$\pm$533.4
& 3296.0$\pm$560.1
& 4289.5$\pm$989.3
& 18734.4$\pm$523.1
& 7826.2$\pm$798.0
& 1353.5$\pm$27.4 \\

\cellcolor{gray!15}{\textbf{w/ AdaConcat}}
& \cellcolor{gray!15}{\textbf{16143.3$\pm$583.0}}
& \cellcolor{gray!15}{\textbf{3688.6$\pm$271.0}}
& \cellcolor{gray!15}{\textbf{6250.2$\pm$757.9}}
& \cellcolor{gray!15}{\textbf{19888.7$\pm$944.6}}
& \cellcolor{gray!15}{\textbf{9670.6$\pm$1176.4}}
& \cellcolor{gray!15}{\textbf{1422.5$\pm$29.5}} \\

\midrule
\midrule

& \textbf{\textsc{Glider-Regular}}& \textbf{\textsc{Glider-Medium}} & \textbf{\textsc{Glider-Hard}} & \textbf{\textsc{Walker-Regular}} & \textbf{\textsc{Walker-Medium}} & \textbf{\textsc{Walker-Hard}} \\
\midrule

w/ Direct Concat
& 20998.1$\pm$1002.0
& 20333.9$\pm$884.1
& 18031.0$\pm$1571.1
& 19197.0$\pm$2119.3
& 19949.6$\pm$856.5
& 20647.6$\pm$851.3 \\

\cellcolor{gray!15}{\textbf{w/ AdaConcat}}
& \cellcolor{gray!15}{\textbf{23317.5$\pm$627.9}}
& \cellcolor{gray!15}{\textbf{22376.4$\pm$809.9}}
& \cellcolor{gray!15}{\textbf{19972.8$\pm$332.1}}
& \cellcolor{gray!15}{\textbf{24306.1$\pm$1293.0}}
& \cellcolor{gray!15}{\textbf{21973.3$\pm$918.9}}
& \cellcolor{gray!15}{\textbf{21614.6$\pm$521.8}} \\

\bottomrule
\end{tabular}
}
\end{table*}

\begin{figure*}[tb]
  \centering
  \includegraphics[width=\linewidth]{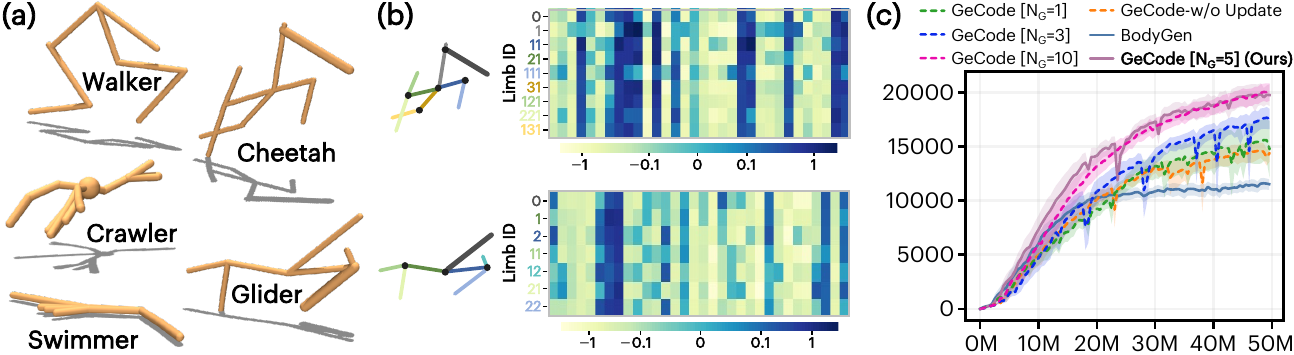}
  \vspace{-0.3in}
  \caption{(a) Representative morphologies designed by GeCode. (b) Visualization of limb-specific conditioning produced by AdaConcat. (c) Ablations on Morphogene updates and set size $N_G$.}
  \label{fig:update}
\end{figure*}

\vspace{-0.05in}
\paragraph{Ablation on AdaConcat}
We evaluate the effectiveness of AdaConcat's adaptive conditioning for Morphogene injection by replacing it with direct concatenation in both the morphology-design and control networks, with results reported in Table~\ref{tab:adaconcat}.
Direct concatenation treats Morphogene as a uniform global feature, disregarding limb-specific context and limiting its capacity to guide specialized design and control decisions. In contrast, AdaConcat adaptively modulates Morphogene for each limb (see Figure~\ref{fig:update}(b)), preserving the shared morphology blueprint while providing context-aware conditioning and thereby enabling finer-grained morphology--control coordination.

\vspace{-0.03in}
\paragraph{Ablation on Morphogene Updates and Set Size}
We isolate the contribution of Morphogene updates in Figure~\ref{fig:update}(c).
Without updates, fixed Morphogenes cannot adapt alongside the evolving morphology-design and control policies, reducing them to static and potentially uninformative conditions. 
This result demonstrates that performance-guided updates are essential for guiding exploration toward promising latent regions and discovering higher-performing designs.

Figure~\ref{fig:update}(c) further examines the effect of Morphogene set size $N_G$. A single Morphogene ($N_G=1$) enables limited exploration but provides insufficient coverage of the design space, whereas too many Morphogenes ($N_G=10$) disperse early rollouts and slow initial convergence. The comparable final performance of $N_G=5$ and $N_G=10$ indicates that five Morphogenes provide sufficient coverage while balancing exploration diversity and learning efficiency.

\begin{figure*}[tb]
  \centering
  \includegraphics[width=\linewidth]{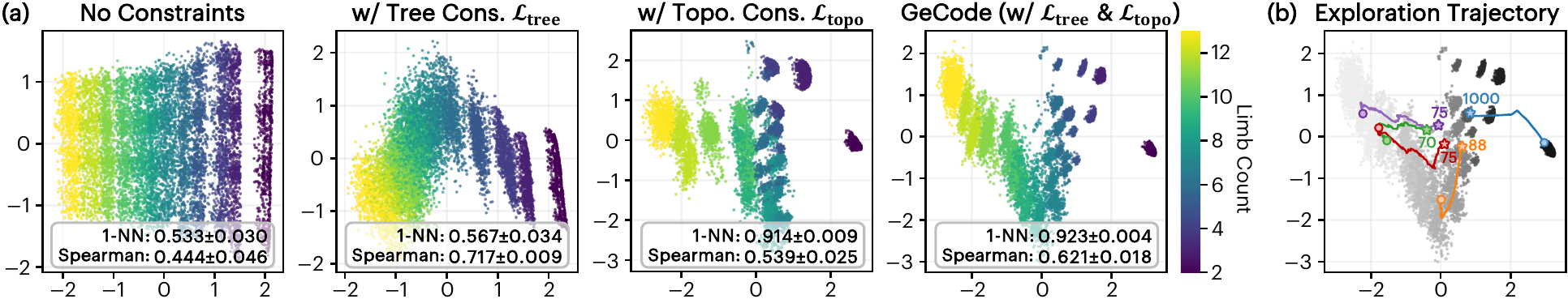}
  \vspace{-0.2in}
  \caption{(a)~Ablation study of structural constraints (i.e., $\mathcal{L}_{\mathrm{topo}}$ and $\mathcal{L}_{\mathrm{tree}}$) in the latent space. We report topology-aware 1-NN consistency and the Spearman rank correlation between latent and tree-edit distances, denoted ``1-NN$\uparrow$'' and ``Spearman$\uparrow$'', respectively.
  (b)~Morphogene update trajectories over 1,000 co-design epochs.
  \mycircle{} and \mystar{} denote the initial and final positions, respectively, while numerical labels indicate elimination epochs. See Figure~\ref{fig:gene_trajectory} in App.~\ref{app:visual} for additional trajectories.}
  \label{fig:latent_space}
\end{figure*}

\paragraph{Ablation on Structural Constraints in the Latent Space}
We further ablate the structural constraints used in RAE training. Figure~\ref{fig:latent_space}(a) visualizes the learned Morphogene spaces and quantifies their local and global organization through topology 1-NN consistency and the Spearman correlation, respectively.
Specifically, $\mathcal{L}_{\mathrm{topo}}$ promotes local clustering among morphologies with shared topologies, whereas $\mathcal{L}_{\mathrm{tree}}$ preserves global structural relationships. By jointly optimizing both objectives, the full model balances local topology consistency with global structural alignment, organizing the latent geometry according to morphological similarity and enabling Morphogene updates to induce smooth, structurally meaningful variations rather than arbitrary perturbations.
Figure~\ref{fig:latent_space}(b) shows that Morphogenes traverse broad regions of the latent space along structurally coherent trajectories.

\begin{figure*}[tb]
\centering
\begin{minipage}[c]{0.36\textwidth}
    \centering
    \includegraphics[width=\linewidth]{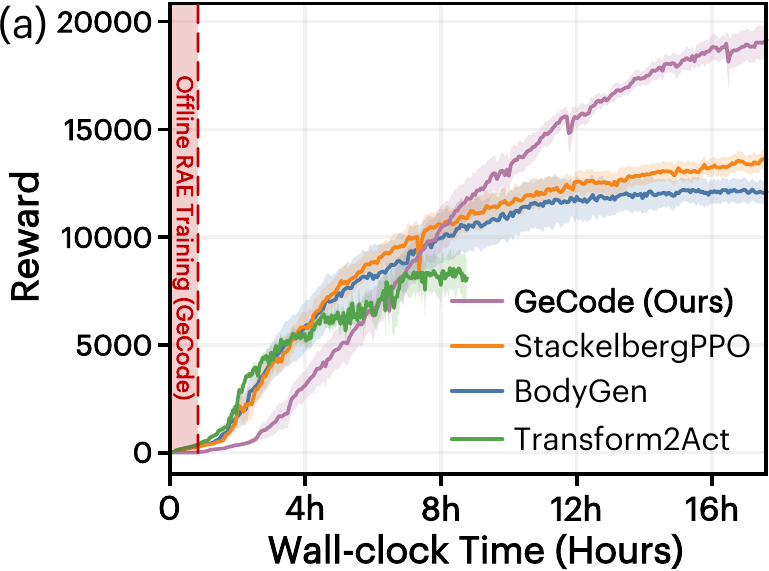}
\end{minipage}
\hfill
\begin{minipage}[c]{0.6\textwidth}
    \centering
    \setlength{\tabcolsep}{0.8 mm}
    \resizebox{\linewidth}{!}{
    \begin{tabular}{@{}lcccc@{}}
    \multicolumn{5}{c}{(b) \textsc{Training-Time Breakdown}} \\
    \toprule
    & \textbf{GeCode} & \textbf{StackelbergPPO} & \textbf{BodyGen} & \textbf{Transform2Act} \\
    \midrule
    Offline RAE Training & 0.77h & N/A   & N/A   & N/A   \\
    Co-design Optimization & 21.38h & 19.96h & 18.02h & 8.94h \\
    \cellcolor{gray!15}{Total Time} & \cellcolor{gray!15}{\textcolor{myred}{22.15h}} & \cellcolor{gray!15}{\textcolor{myyellow}{19.96h}} & \cellcolor{gray!15}{\textcolor{myblue}{18.02h}} & \cellcolor{gray!15}{\textcolor{mygreen}{8.94h}} \\
    \bottomrule
    \noalign{\vskip 1.5ex}
    \multicolumn{5}{c}{(c) \textsc{Cost to Match Baseline Performance}} \\
    \toprule
    & & \textbf{StackelbergPPO} & \textbf{BodyGen} & \textbf{Transform2Act} \\
    \midrule
    \multirow{2}{*}{Wall-clock Time}  & \textbf{GeCode} & \textcolor{myyellow}{10.30h} & \textcolor{myblue}{9.18h} & \textcolor{mygreen}{7.06h} \\
    & & \textcolor{red}{$\downarrow$ \textbf{9.66h}} & \textcolor{red}{$\downarrow$ \textbf{8.84h}} & \textcolor{red}{$\downarrow$ \textbf{1.88h}} \\
    \midrule
    \multirow{2}{*}{\shortstack{Environment Interactions\\(50M Total)}} & \textbf{GeCode} & 21.2M & 18.7M & 13.7M \\
    & & \textcolor{red}{$\downarrow$ \textbf{28.8M}} & \textcolor{red}{$\downarrow$ \textbf{31.3M}} & \textcolor{red}{$\downarrow$ \textbf{36.3M}} \\
    \bottomrule
    \end{tabular}
    }
\end{minipage}
\vspace{-0.1in}
\caption{Computational efficiency of GeCode. 
(a)~Performance comparison under a wall-clock time budget. 
(b)~Detailed training costs, together with the resources required for GeCode to match each baseline's best performance. Further results are provided in Table~\ref{tab:app_wall_clock} and Table~\ref{tab:inference_cost} in App.~\ref{sec:app_efficient}.}
\label{fig:wall_clock}
\end{figure*}

\vspace{-0.1in}
\paragraph{Analysis of Computational Cost}
We evaluate computational efficiency against wall-clock time.
As shown in Figure~\ref{fig:wall_clock}(a), multiple Morphogenes expose the controller to diverse morphologies early in training, imposing a modest initial adaptation cost. As performance-guided updates focus the anchors on promising morphologies and compatible controllers, GeCode converges rapidly and substantially outperforms the baselines under matched time budgets.
Its significantly higher final returns further indicate superior body--brain designs.
Specifically, GeCode reaches StackelbergPPO's peak performance with 48.4\% less wall-clock time and 28.8M fewer environment interactions (Figure~\ref{fig:wall_clock}(c)).
Meanwhile, one-time offline RAE training requires no environment interaction and accounts for only 3.5\% of total training time, while Morphogene updates reuse existing rollouts and require only lightweight latent operations, without additional simulations or optimization loops.

\section{Conclusion}
We introduced GeCode, a Morphogene-driven framework for morphology--control co-design. 
Morphogene bridges body and brain by jointly conditioning morphology generation and controller adaptation through AdaConcat.
GeCode further reformulates co-design as exploration in the compact Morphogene space, where Morphogenes serve as searchable design anchors. 
Stochastic rollouts enable local exploration around each anchor, while performance-guided updates move the anchors toward promising regions for global exploration.
Extensive experiments show that GeCode achieves faster convergence and higher final performance. 
Limitations are discussed in App.~\ref{app:limit}.

\subsubsection*{Acknowledgments}
We sincerely thank Yanning Dai from KAUST for the extensive discussions and valuable assistance with experimental design, experimental implementation, and result analysis.
This research was supported by the Jiangsu Science Foundation (BG2024036, BK20243012), the National Natural Science Foundation of China (625B2045, 62125602, U24A20324, 92464301, 62306073), the New Cornerstone Science Foundation through the XPLORER PRIZE, the Fundamental Research Funds for the Central Universities (2242025K30024), and SEU Innovation Capability Enhancement Plan for Doctoral Students (CXJH\_SEU 26023).

\subsection*{AI use statement}

The authors used generative AI tools to assist with writing, LaTeX and figure formatting, and code editing. All claims, citations, code, figures, and numerical results were independently reviewed and verified by the authors, who take full responsibility for the final submission.

\subsection*{Ethics statement}

This work studies morphology--control co-design in simulated environments using open-source simulators and benchmarks. It involves no human participants, animals, personal or sensitive data. We identify no specific ethical risks associated with the experiments presented in this work.

\subsection*{Reproducibility statement}
We describe GeCode and the experimental protocol in detail in the main paper and appendix. The appendix further provides environment and task specifications, algorithmic and implementation details, and analyses of sample and training efficiency. Additional results and ablation studies are included to facilitate reproducibility.

\bibliography{iclr2027_conference}
\bibliographystyle{iclr2027_conference}

\clearpage
\appendix
\section{Additional Related Work}
\label{app:related}
\paragraph{``Genes'' in Intelligent Agents}
Biological genes have inspired diverse machine learning methods.
In evolutionary search for intelligent agents, candidate architectures and morphologies are encoded as gene-like genotypes, often binary strings, and evolved through selection and variation~\citep{gupta2021embodied,qiu2025robomorph,stanley2019designing,marchesini2020genetic}. Such genotypes serve as search variables rather than semantically meaningful agent-level abstractions.
Unlike search-oriented genotypes, learngenes~\citep{feng2023genes} focus on knowledge inheritance by encoding task-agnostic knowledge from ancestral models into compact, inheritable network fragments for efficient downstream adaptation~\citep{feng2026knowledge,xie2025kind,xie2025divcontrol}.
Unlike both, Morphogene is an agent-level blueprint that jointly guides morphology and control generation, serving as a bridge for body--brain coordination rather than a search genotype or a carrier of transferable knowledge.

\paragraph{Universal Morphology Control}
Universal morphology control aims to learn a single policy capable of controlling agents with diverse body structures, thereby avoiding separate controller training for each morphology~\citep{nagabandilearning,pathak2019learning,patel2025get}. Early MLP-based approaches have limited flexibility when agents differ in limb count and action dimensionality~\citep{ghadirzadeh2021bayesian,feng2023genloco}. Graph neural networks address this limitation by representing limbs or actuators as nodes and propagating information according to morphological topology~\citep{wang2018nervenet,huang2020one}. More recently, MetaMorph employs a morphology-aware Transformer to model whole-body interactions~\citep{guptametamorph}, whereas ModuMorph~\citep{xiong2023universal} and HyperDistill~\citep{xiong2024distilling} improve cross-morphology generalization through morphology-conditioned modulation and policy distillation, respectively. 
DivMorph~\citep{feng2026knowledge} further adopts a modular training paradigm that leverages knowledge diversion~\citep{xie2025kind} to learn decomposable Transformer controllers, preserving agent-specific knowledge while supporting unified control and transfer. Collectively, these advances establish an important foundation for our Transformer-based approach to morphology--control co-design.

\section{Environment and Task Details}
\label{app:env}

\subsection{Benchmark Setup}
\begin{figure*}[p]
  \centering
  \includegraphics[width=\linewidth]{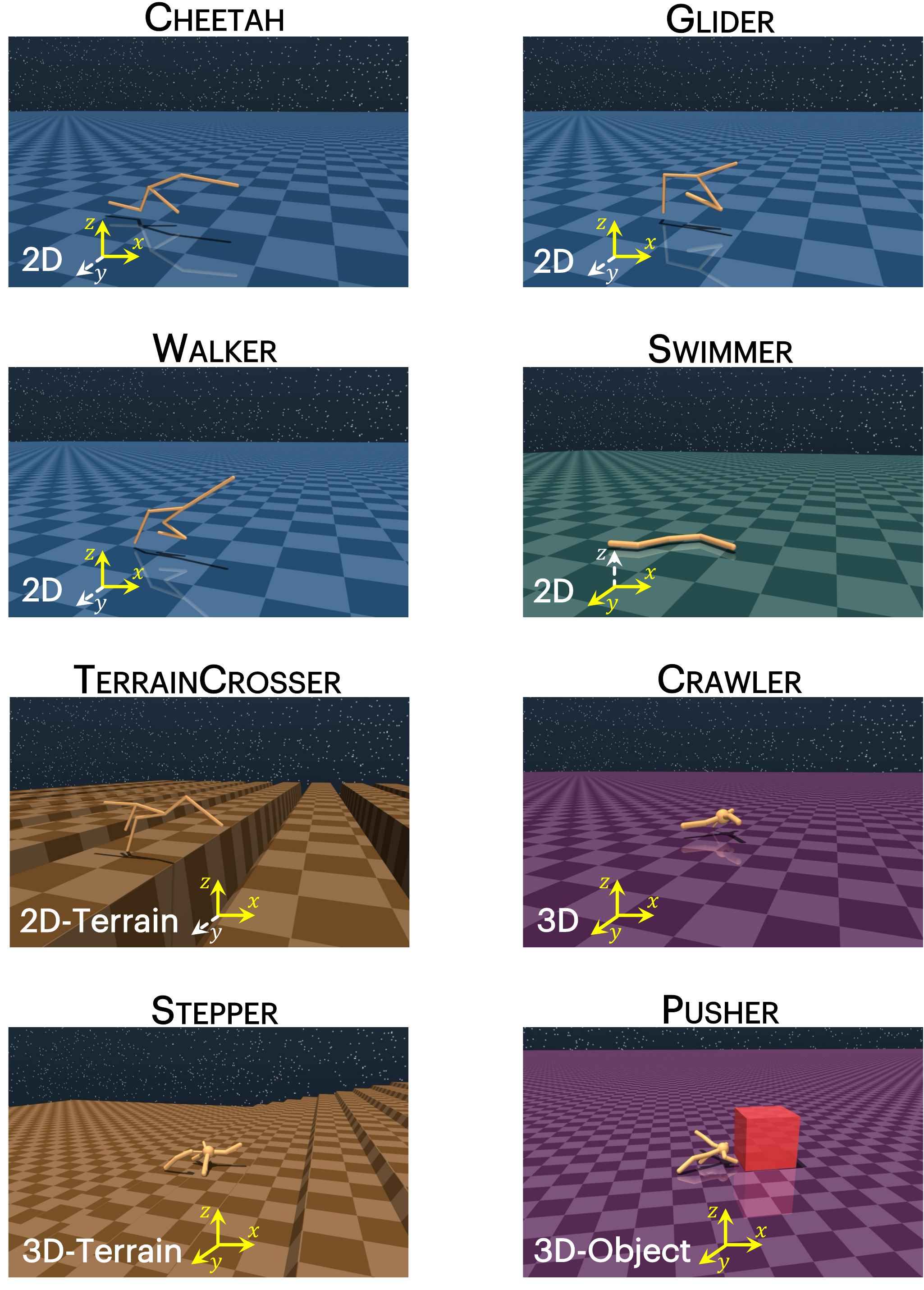}
  \vspace{-0.3in}
  \caption{\textbf{Overview of the Benchmark Environments.}
  \textsc{Crawler}, \textsc{Stepper}, and \textsc{Pusher} are three-dimensional (3D) tasks, whereas the remaining environments are two-dimensional (2D), operating in either the $x$--$z$ or $x$--$y$ plane. 
  The benchmarks span diverse morphology depths, symmetries, limb configurations, terrains, and task objectives, enabling a comprehensive evaluation of morphology--control co-design across locomotion and manipulation settings.}
  \label{fig:visual_env}
  \vspace{-0.1in}
\end{figure*}

\paragraph{Benchmark Environments} Following prior work~\citep{lu2025bodygen,dai2026stackelberg}, we evaluate GeCode on twelve morphology--control co-design tasks: \textsc{Crawler}, \textsc{Stepper}, \textsc{Pusher}, \textsc{TerrainCrosser}, \textsc{Cheetah}, \textsc{Swimmer}, \textsc{Glider-(Regular, Medium, Hard)}, and \textsc{Walker-(Regular, Medium, Hard)}. Figure~\ref{fig:visual_env} provides a visual overview of the corresponding task environments.
These tasks span planar and three-dimensional locomotion, uneven-terrain traversal, and object manipulation, providing a diverse benchmark for evaluating the effectiveness of morphology--control co-design across morphology spaces and task complexities.

Each agent is modeled as a morphology graph $\mathcal{M}=(V,E,A)$, where each node $u\in V$ represents a rigid limb and each edge $e{(u,v)}\in E$ represents a motor-actuated joint. The limb attributes $\Lambda_u^{V}$ specify geometric properties, such as length and size, while the joint attributes $\Lambda_{e{(u,v)}}^{E}$ specify mechanical properties, such as rotation limits and maximum motor torque. 

\vspace{-0.1in}
\begin{figure*}[ht]
  \centering
  \includegraphics[width=0.85\linewidth]{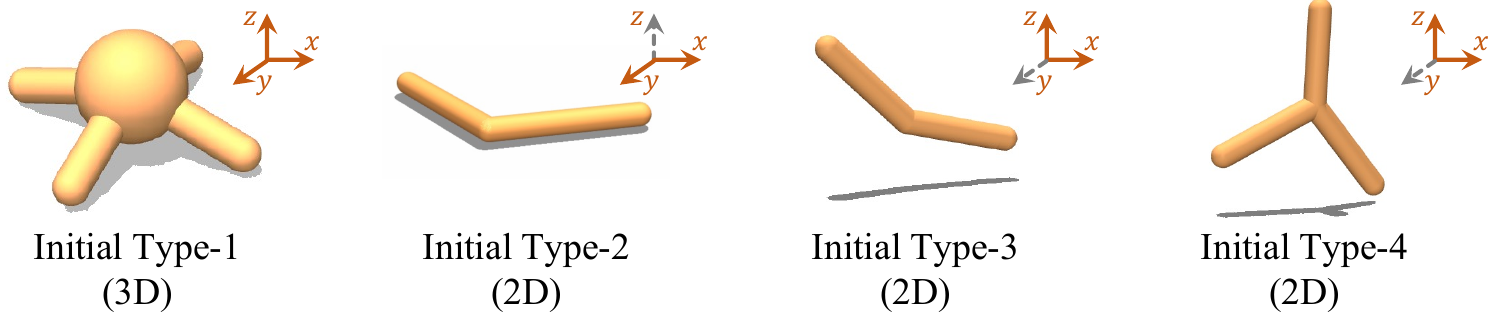}
  \vspace{-0.1in}
  \caption{\textbf{Overview of Task-Specific Initial Morphologies.}
  Initial Type~1 is a three-dimensional morphology comprising four limbs. 
  Initial Type~2 comprises two limbs connected by one joint in the $x$--$y$ plane. 
  Initial Type~3 comprises two limbs connected by one joint in the $x$--$z$ plane. 
  Initial Type~4 comprises three limbs connected by two joints in the $x$--$z$ plane. These morphologies serve as the starting points for sequential morphology design.}
  \label{fig:initial}
\end{figure*}

\paragraph{Co-Design Procedure} Each episode comprises a morphology-design stage $\Phi_{\mathrm{des}}$ followed by a control stage $\Phi_{\mathrm{ctrl}}$. 
Morphology design begins from a task-specific initial morphology $\mathcal{M}_0$, selected from the four structures shown in Figure~\ref{fig:initial}. 
Over $T_{\mathrm{des}}$ steps, the design policy $\pi_{\phi}^{\mathrm{des}}$ sequentially updates $\mathcal{M}_t$ through morphology-editing actions $a_t^{\mathrm{des}}$, first modifying the discrete topology and then updating continuous physical attributes. During topology design, $a_t^{\mathrm{des}}$ may add or remove a terminal limb or leave the structure unchanged. Each edit must satisfy task-specific upper bounds on tree depth, total limb count, and the number of children per limb, ensuring structurally valid morphologies. The subsequent attribute update specifies limb offsets, capsule radii, attachment positions, and actuator gear ratios. 
The resulting final morphology $\mathcal{M}^{\star}=\mathcal{M}_{T_{\mathrm{des}}}$ remains fixed throughout $\Phi_{\mathrm{ctrl}}$.

Once the design stage $\Phi_{\mathrm{des}}$ is complete, the simulator instantiates the final morphology $\mathcal{M}^{\star}$ as a physical agent and fixes its structural and physical parameters throughout the control stage $\Phi_{\mathrm{ctrl}}$. 
At each control step $t$, the policy $\pi_{\theta}^{\mathrm{ctrl}}$ maps the proprioceptive state $s_t^{\mathrm{ctrl}}$, which includes position, linear velocity, and angular velocity, to a continuous motor action $a_t^{\mathrm{ctrl}}$.
The resulting cumulative task return evaluates the body--brain pair $(\mathcal{M}^{\star},\pi_{\theta}^{\mathrm{ctrl}})$ and provides the shared objective for jointly optimizing the design policy $\pi_{\phi}^{\mathrm{des}}$ and control policy $\pi_{\theta}^{\mathrm{ctrl}}$.

\subsection{Task-Specific Specifications}
We next describe the environment, objective, and design constraints of each task.

\paragraph{Crawler}
\textsc{Crawler} evaluates three-dimensional forward locomotion over level terrain. 
Its initial morphology is the Type~1 initial morphology in Figure~\ref{fig:initial}, comprising a free-floating root body and four primary limbs. 
The morphology is limited to a tree depth of 4, a total of 29 bodies, and at most two child limbs per body. During attribute design, the root offset remains fixed, while non-root limb offsets are optimized in the $x$--$y$ plane within $[-0.5,0.5]^2$. Capsule radii, attachment positions, and actuator gear ratios are constrained to $[0.03,0.10]$, $[0,0.2]$, and $[20,400]$, respectively.

The reward balances forward velocity and control effort:
\begin{equation}
r_t
=
\frac{x_{t+1}-x_t}{\Delta t}
-
\omega\cdot \frac{1}{N_L}
\left\|a_t^{\mathrm{ctrl}}\right\|_2^2,
\label{equ:reward_3D}
\end{equation}
where $x_t$ denotes the root position along the forward axis, $\Delta t=0.04$ is the control interval, $\omega=10^{-4}$ weights the control penalty, $N_L$ is the number of limbs, and $a_t^{\mathrm{ctrl}}$ is the continuous motor-action vector. An episode terminates if the root height falls outside $(0,2.0)$ or its tilt from the upright orientation exceeds $60^\circ$.

\paragraph{Stepper}
\textsc{Stepper} evaluates three-dimensional locomotion over a sequence of elevated steps. It uses the Type~1 initial morphology in Figure~\ref{fig:initial} and shares the topology constraints and continuous attribute ranges of \textsc{Crawler}: a maximum tree depth of 4, at most 29 bodies, and no more than two child limbs per body. The task adopts the forward-velocity and control-effort reward in Eq.~(\ref{equ:reward_3D}), with control interval $\Delta t=0.04$ and penalty weight $\omega=10^{-4}$.
Because the terrain elevation varies along the course, termination is determined by the root height relative to the local ground surface. An episode terminates if this relative height falls outside $(0,2.0)$ or if the root tilt from the upright orientation exceeds $90^\circ$. The task therefore favors morphologies with sufficient support, ground clearance, and reach to maintain forward progress across abrupt elevation changes.

\paragraph{Pusher}
\textsc{Pusher} evaluates three-dimensional object manipulation using the Type~1 initial morphology shown in Figure~\ref{fig:initial}.
The task requires the agent to push a movable cube along the positive $x$-axis while minimizing lateral drift along the $y$-axis. The morphology is constrained to a maximum tree depth of 4, at most 29 bodies, and no more than two child limbs per body. Its continuous attribute ranges are identical to those of \textsc{Crawler} and \textsc{Stepper}.

Let $(x_t^{\mathrm{cube}},y_t^{\mathrm{cube}})$ denote the planar coordinates of the cube, and let $\mathbf{p}_t^{\mathrm{cube}}$ and $\mathbf{p}_t^{\mathrm{root}}$ denote the three-dimensional positions of the cube and robot root, respectively. The reward is
\begin{equation}
r_t
=
\frac{x_{t+1}^{\mathrm{cube}}-x_t^{\mathrm{cube}}}{\Delta t}
-
\kappa
\frac{|y_{t+1}^{\mathrm{cube}}-y_t^{\mathrm{cube}}|}{\Delta t}
+
\frac{1}{
1+\|\mathbf{p}_{t+1}^{\mathrm{cube}}-\mathbf{p}_{t+1}^{\mathrm{root}}\|_2
}
-
\omega\cdot\frac{1}{N_L}
\left\|a_t^{\mathrm{ctrl}}\right\|_2^2.
\label{equ:reward_pusher}
\end{equation}
Here, $\kappa=0.1$ weights lateral drift, $\Delta t=0.04$ is the control interval, and $\omega=10^{-4}$ weights control effort. The four terms reward forward cube motion, penalize lateral drift, encourage robot--cube proximity, and penalize control effort, respectively. An episode terminates if the root height falls outside $(0,2.0)$ or its tilt from the upright orientation exceeds $60^\circ$.

\paragraph{Cheetah}
\textsc{Cheetah} evaluates rapid planar locomotion along the positive $x$-axis. The agent uses the Type~3 initial morphology in Figure~\ref{fig:initial}, comprising a root body and one primary limb arranged in the $x$--$z$ plane. The morphology is constrained to a maximum tree depth of 4, at most 14 bodies, and no more than three child limbs per body. Non-root limb offsets are optimized within $[-0.5,0.5]^2$ in the $x$--$z$ plane, while capsule radii, attachment positions, and actuator gear ratios are restricted to $[0.03,0.08]$, $[0,0.5]$, and $[20,400]$, respectively.

The reward is defined by the forward velocity of the root:
\begin{equation}
r_t
=
\frac{x_{t+1}-x_t}{\Delta t}.
\label{equ:reward_cheetah}
\end{equation}
Here, $x_t$ denotes the root position along the forward axis and $\Delta t=0.008$ is the control interval. 
An episode terminates if the root height falls outside $(0.7,2.0)$ or if its absolute angular deviation from the upright orientation exceeds $20^\circ$.

\paragraph{TerrainCrosser}
\textsc{TerrainCrosser} evaluates planar locomotion over discontinuous terrain with repeated gaps. It uses the Type~3 initial morphology in Figure~\ref{fig:initial} and shares the topology constraints and continuous attribute ranges of \textsc{Cheetah}: a maximum tree depth of 4, at most 14 bodies, and no more than three child limbs per body. The task adopts the forward-velocity reward in Eq.~(\ref{equ:reward_cheetah}), with control interval $\Delta t=0.008$.
Compared with \textsc{Cheetah}, \textsc{TerrainCrosser} imposes a stricter minimum root height. An episode terminates if the root height falls outside $(1.0,2.0)$ or if the absolute root pitch exceeds $20^\circ$. The task therefore favors morphologies that combine rapid forward movement with sufficient reach, clearance, and stability to cross gaps.

\paragraph{Swimmer}
\textsc{Swimmer} evaluates planar propulsion through a viscous fluid. It uses the Type~2 initial morphology in Figure~\ref{fig:initial}, comprising a root segment connected to one actuated limb in the $x$--$y$ plane. The morphology is constrained to a maximum tree depth of 4, at most 14 bodies, and no more than three child limbs per body. Non-root limb offsets are optimized within $[-1.5,1.5]^2$, while capsule radii, attachment positions, and actuator gear ratios are restricted to $[0.06,0.20]$, $[0,0.2]$, and $[20,300]$, respectively.
The task adopts the forward-velocity and control-effort reward in Eq.~(\ref{equ:reward_3D}), with control interval $\Delta t=0.04$ and penalty weight $\omega=10^{-4}$. No posture-based early termination is applied; each episode continues until the predefined time horizon.

\paragraph{Glider-Regular}
\textsc{Glider-Regular} evaluates planar locomotion within a compact, weakly branching morphology space. It uses the Type~4 initial morphology in Figure~\ref{fig:initial}, comprising a root body and two primary limbs. The morphology is restricted to a maximum tree depth of 3, at most 5 bodies, and no more than one child limb per body, favoring chain-like structures. Limb offsets are optimized within $[-0.5,0.5]^2$ in the task plane, while capsule radii, attachment positions, and actuator gear ratios are constrained to $[0.03,0.08]$, $[0,0.5]$, and $[20,400]$, respectively.
The task adopts the forward-velocity reward in Eq.~(\ref{equ:reward_cheetah}), with control interval $\Delta t=0.008$. An episode terminates if the root height falls outside $(0.7,2.0)$ or if the absolute root pitch exceeds $60^\circ$.

\paragraph{Glider-Medium}
\textsc{Glider-Medium} uses the same Type~4 initial morphology, environment dynamics, continuous attribute ranges, and termination conditions as \textsc{Glider-Regular}. It also adopts the forward-velocity reward in Eq.~(\ref{equ:reward_cheetah}), with $\Delta t=0.008$. The maximum tree depth remains 3, while the body limit increases from 5 to 7 and the maximum number of child limbs per body increases from one to two. This larger branching factor introduces a moderately more complex topology space.

\paragraph{Glider-Hard}
\textsc{Glider-Hard} further expands the topology space by increasing the body limit from 7 to 9 and allowing up to three child limbs per body, while retaining a maximum tree depth of 3.
All other design ranges, environment dynamics, reward, and termination conditions remain unchanged.
Together, the three \textsc{Glider} settings provide a controlled evaluation of increasing morphology size and branching complexity under an otherwise identical locomotion objective.

\paragraph{Walker-Regular}
\textsc{Walker-Regular} evaluates planar locomotion in a deeper morphology space than the \textsc{Glider} tasks, starting from the Type~4 morphology shown in Figure~\ref{fig:initial}. The design space permits a maximum tree depth of 4, at most 7 bodies, and no more than one child limb per body, thereby favoring deep, chain-like structures. Limb offsets are optimized within $[-0.5,0.5]^2$ in the task plane, while capsule radii, attachment positions, and actuator gear ratios are restricted to $[0.03,0.08]$, $[0,0.5]$, and $[20,400]$, respectively. The task uses the forward-velocity reward defined in Eq.~(\ref{equ:reward_cheetah}), with a control interval of $\Delta t=0.008$. An episode terminates if the root height falls outside $(0.7,2.0)$ or if the absolute root pitch exceeds $60^\circ$.

\paragraph{Walker-Medium}
\textsc{Walker-Medium} retains the initial morphology, environment dynamics, continuous attribute ranges, reward, and termination conditions of \textsc{Walker-Regular}. The maximum tree depth remains 4, while the body limit increases from 7 to 15 and the maximum number of child limbs per body increases from one to two. This expanded topology space accommodates both chain-like and moderately branched locomotor structures.

\paragraph{Walker-Hard}
\textsc{Walker-Hard} further expands the topology space by increasing the body limit from 15 to 27 and allowing up to three child limbs per body, while retaining a maximum tree depth of 4. The initial morphology, environment dynamics, continuous attribute ranges, reward, and termination conditions remain unchanged. This setting defines the largest planar design space among the \textsc{Walker} tasks and requires coordinated control of a potentially larger and more highly branched articulated body.

\section{Method Details}
\label{app:method_detail}
\subsection{Algorithm Details}
GeCode extends BodyGen~\citep{lu2025bodygen} with a compact Morphogene space for explicit coordination between morphology design and control. The overall procedure comprises two phases. First, a structure-aware RAE is trained offline on randomly generated morphologies to construct the Morphogene space. GeCode then performs Morphogene-driven co-design using a small set of Morphogenes as latent design anchors, which jointly condition the morphology-design and control policies through AdaConcat. During co-design, policy rollouts provide both learning experience and performance statistics for iteratively updating, eliminating, and resampling these anchors. The complete procedure is summarized in Algorithm~\ref{alg:gecode}.

\begin{algorithm}[ht]
\caption{GeCode: Morphogene-Driven Morphology--Control Co-Design}
\label{alg:gecode}
\DontPrintSemicolon

\KwIn{Randomly initialized morphology-design and control policies
$\pi_{\phi}^{\mathrm{des}}$ and $\pi_{\theta}^{\mathrm{ctrl}}$;
Morphogene set size $N_G$;
Morphogene-space construction epochs $E_{\mathrm{I}}$;
Morphogene-driven co-design epochs $E_{\mathrm{II}}$.}

\BlankLine
{%
\renewcommand{\nl}{}%
\setlength{\fboxsep}{1pt}%
\colorbox{gray!15}{%
    \parbox[t]{\dimexpr0.95\linewidth-2\fboxsep\relax}{%
        \textbf{\textsc{Stage} I: \textsc{Morphogene-space Construction}}%
    }%
}\;
}

Randomly generate and instantiate a collection of morphologies
$\mathcal{D}=\{\mathcal{M}_j^{\mathrm{vec}}\}_{j=1}^{N_D}$\;
\For{$e=1,\ldots,E_{\mathrm{I}}$}{
    Sample a morphology mini-batch $\mathcal{B}\subset\mathcal{D}$\;
    Encode $\mathcal{M}_j^{\mathrm{vec}}\in\mathcal{B}$ as
    $\mathbf{z}_j^{\mathrm{morph}}=E_{\psi}(\mathcal{M}_j^{\mathrm{vec}})$
    using Eq.~(\ref{eq:morphogene-encoder})\;
    Reconstruct
    $(\widehat{\mathcal{M}}_j^{\mathrm{vec}},\widehat{\boldsymbol{\chi}}_j)
    =D_{\psi}(\mathbf{z}_j^{\mathrm{morph}})$
    using Eq.~(\ref{eq:morphogene-decoder})\;
    Update $\psi$ with reconstruction loss in Eq.~(\ref{eq:rae-objective}) and structural losses in Eqs.~(\ref{eq:topology_contrastive}) and~(\ref{eq:tree_metric})\;
}
Retain and freeze $E_{\psi}$ for co-design, and discard $D_{\psi}$\;
Independently sample
$\mathcal{P}_{\mathcal{G}}=\{\mathcal{G}_i\}_{i=1}^{N_G}$
from the learned latent space\;
\textbf{\textit{Initialize}} sampling probabilities $p_i\leftarrow \frac{1}{N_G}$\;

\BlankLine
{%
\renewcommand{\nl}{}%
\setlength{\fboxsep}{1pt}%
\colorbox{gray!15}{%
    \parbox[t]{\dimexpr0.95\linewidth-2\fboxsep\relax}{%
        \textbf{\textsc{Stage} II: \textsc{Morphogene-driven Co-design}}%
    }%
}\;
}
\For{$e=1,\ldots,E_{\mathrm{II}}$}{
    \For{each rollout $k$}{
        Sample a Morphogene $\mathcal{G}_i\in\mathcal{P}_{\mathcal{G}}$ according to $i\sim\mathbf{p}$\;
        Generate $\mathcal{M}_{i,k}$ using the Morphogene-conditioned design policy $\pi_{\phi}^{\mathrm{des}}(\cdot\mid\mathcal{G}_i)$ with AdaConcat, following Eqs.~(\ref{equ:adaconcat}) and~(\ref{equ:gene_action})\;
        Instantiate and encode the morphology as
        $\mathbf{z}_{i,k}^{\mathrm{morph}}=E_{\psi}(\mathcal{M}_{i,k}^{\mathrm{vec}})$ using Eq.~(\ref{eq:morphogene-encoder})\;
        Compute the terminal proximity reward $r_t^{\mathrm{prox}}$ using Eq.~(\ref{equ:proximity_reward})\;
        Execute the control policy $\pi_{\theta}^{\mathrm{ctrl}}(\cdot\mid\mathcal{G}_i)$ on $\mathcal{M}_{i,k}$, conditioned on the same Morphogene through AdaConcat according to Eqs.~(\ref{equ:adaconcat}) and~(\ref{equ:gene_action})\;
        Record the trajectory and task return $R_{i,k}$\;
    }
    \textbf{\textit{Optimize}} $(\phi,\theta)$ using collected trajectories
    under Eq.~(\ref{equ:joint_objective})\;
    Construct local hit sets
    $\mathcal{K}_i=\{k\mid
    \operatorname{sim}(\mathbf{z}_{i,k}^{\mathrm{morph}},
    \mathcal{G}_i)\geq\tau_{\mathrm{sim}}\}$
    and their union
    $\mathcal{K}=\{(i,k)\mid k\in\mathcal{K}_i\}$\;
    \If{$\mathcal{K}\neq\varnothing$}{
        Select \textit{global-best} code $\mathbf{z}^{\mathrm{glob}}$
        by maximizing $R_{i,k}$ over $(i,k)\in\mathcal{K}$\;
        \ForEach{active $\mathcal{G}_i$ with
        $\mathcal{K}_i\neq\varnothing$}{
            Select \textit{local-best} code $\mathbf{z}_i^{\mathrm{loc}}$
            by maximizing $R_{i,k}$ over $k\in\mathcal{K}_i$\;
            \textbf{\textit{Update}} each Morphogene $\mathcal{G}_i\in\mathcal{P}_{\mathcal{G}}$ according to Eq.~(\ref{equ:update})\;
        }
    }
    \textbf{\textit{Eliminate}} each Morphogene $\mathcal{G}_i\in\mathcal{P}_{\mathcal{G}}$ with $|\mathcal{K}_i|<H_{\min}$ (Eq.~\ref{equ:eliminate})\;
    \textbf{\textit{Update}} the sampling distribution $\mathbf{p}$ over the retained Morphogenes using Eq.~(\ref{equ:update_P})\;
}
\end{algorithm}

\subsection{Structural Regularization of the Morphogene Space}
\label{app:morphogene_structural_losses}

As discussed in Sec.~\ref{sec:morphgene}, reconstruction preserves the information required to recover a morphology but does not explicitly organize the latent geometry. We therefore impose two complementary structural regularizers on the morphology codes: a topology-aware contrastive loss $\mathcal{L}_{\mathrm{topo}}$ for local topology clustering and a tree-structured metric loss $\mathcal{L}_{\mathrm{tree}}$ for global structural alignment.

For a mini-batch $\mathcal{B}$, let
$\mathbf{z}_i^{\mathrm{morph}}=E_{\psi}(\mathcal{M}_i^{\mathrm{vec}})$
denote the morphology code of sample $i$, and let
$\widetilde{\mathbf{z}}_i^{\mathrm{morph}}
=
\frac{\mathbf{z}_i^{\mathrm{morph}}}
{\|\mathbf{z}_i^{\mathrm{morph}}\|_2}$
be its $\ell_2$-normalized representation. The topology-aware contrastive loss treats morphologies with isomorphic topology graphs as positive pairs:
\begin{equation}
\mathcal{L}_{\mathrm{topo}}
=
-\frac{1}{|\mathcal{B}_{+}|}
\sum_{i\in\mathcal{B}_{+}}
\frac{1}{|\mathcal{P}(i)|}
\sum_{p\in\mathcal{P}(i)}
\log
\frac{
\exp\!\left(
(\widetilde{\mathbf{z}}_{i}^{\mathrm{morph}})^{\top}
\widetilde{\mathbf{z}}_{p}^{\mathrm{morph}}
\big/\tau_{\mathrm{topo}}
\right)
}{
\displaystyle
\sum_{k\in\mathcal{B}\setminus\{i\}}
\exp\!\left(
(\widetilde{\mathbf{z}}_{i}^{\mathrm{morph}})^{\top}
\widetilde{\mathbf{z}}_{k}^{\mathrm{morph}}
\big/\tau_{\mathrm{topo}}
\right)
}.
\label{eq:topology_contrastive}
\end{equation}
Here,
$\mathcal{P}(i)
=
\{p\in\mathcal{B}\setminus\{i\}\mid
\mathcal{T}_p\cong\mathcal{T}_i\}$
contains the samples whose topology graphs are isomorphic to that of sample $i$,
$\mathcal{B}_{+}
=
\{i\in\mathcal{B}\mid|\mathcal{P}(i)|>0\}$
contains samples with at least one positive pair, and
$\tau_{\mathrm{topo}}$ is the temperature parameter. This objective draws morphology codes with the same topology closer while separating those with different topologies.

To preserve finer-grained structural relationships beyond topology classes, we construct a set of index triplets
$(i,i_{+},i_{-})\in\mathcal{Q}$, corresponding to a reference morphology, a structurally similar morphology, and a structurally dissimilar morphology, respectively. The positive index $i_{+}$ is selected based on a small tree-edit distance from $i$, whereas the negative index $i_{-}$ is selected based on a large tree-edit distance. The tree-structured metric loss enforces the corresponding reference code to be closer to the positive code than to the negative code by a margin:
\begin{equation}
\mathcal{L}_{\mathrm{tree}}
=
\frac{1}{|\mathcal{Q}|}
\sum_{(i,i_{+},i_{-})\in\mathcal{Q}}
\max\!\left(
0,\,
d_{\cos}\!\left(
\widetilde{\mathbf{z}}_{i}^{\mathrm{morph}},
\widetilde{\mathbf{z}}_{i_{+}}^{\mathrm{morph}}
\right)
-
d_{\cos}\!\left(
\widetilde{\mathbf{z}}_{i}^{\mathrm{morph}},
\widetilde{\mathbf{z}}_{i_{-}}^{\mathrm{morph}}
\right)
+
\delta_{\mathrm{tree}}
\right).
\label{eq:tree_metric}
\end{equation}
Here, $d_{\cos}(\mathbf{u},\mathbf{v})=1-\mathbf{u}^{\top}\mathbf{v}$ denotes the cosine distance between normalized morphology codes, and $\delta_{\mathrm{tree}}$ is the ranking margin.

The complete RAE training objective is
\begin{equation}
\mathcal{L}_{\mathrm{RAE}}
=
\mathcal{L}_{\mathrm{rec}}
+
\lambda_{\mathrm{topo}}\mathcal{L}_{\mathrm{topo}}
+
\lambda_{\mathrm{tree}}\mathcal{L}_{\mathrm{tree}},
\end{equation}
where $\lambda_{\mathrm{topo}}$ and $\lambda_{\mathrm{tree}}$ control the contributions of the two structural losses.

\subsection{Offline and RL-Based Training of the Morphology Decoder}

In this work, we learn the morphology latent space using an offline, reconstruction-based RAE decoder, as described in Sec.~\ref{sec:morphgene}. As an alternative, the morphology-design policy can serve as an RL-trained decoder: rather than reconstructing morphology attributes directly, it maps a target morphology code to a morphology through sequential design actions. 
We describe this alternative below and discuss its trade-offs relative to offline reconstruction.

\paragraph{RL-based Policy Decoding}
The morphology-design policy $\pi_\phi^{\mathrm{des}}$ can be trained to decode a target morphology code $\mathbf{z}^{\mathrm{morph}}$ into a valid design. Starting from an initial morphology $\mathcal{M}_0$, the policy generates topology and attribute design actions conditioned on the current morphology and target code, $a_t^{\mathrm{des}} \sim \pi_\phi^{\mathrm{des}}(\cdot \mid \mathcal{M}_t; \mathbf{z}^{\mathrm{morph}})$, yielding the final morphology $\mathcal{M}^\star=\mathcal{M}_{T_{\mathrm{des}}}$. We train the policy using the proximity reward in Eq.~(\ref{equ:proximity_reward}), with the target Morphogene $\mathcal{G}$ set to $\mathbf{z}^{\mathrm{morph}}$. This reward encourages the code of the generated morphology to meet the prescribed similarity threshold with respect to the target code.

RL-based decoding requires simulator rollouts and policy optimization, making pretraining more computationally demanding than direct supervised reconstruction. Moreover, the proximity reward alone does not explicitly impose the pairwise topological and tree-structural relationships encouraged by the regularizers in Eqs.~(\ref{eq:topology_contrastive}) and~(\ref{eq:tree_metric}); consequently, it provides less direct control over the organization of the latent space. Its principal advantage is that pretraining also optimizes the morphology-design policy. The pretrained policy can then be carried into co-design, where it continues to generate morphologies while being optimized jointly with the control policy. 
Accordingly, RL-based decoding incurs higher pretraining cost than offline reconstruction and, without additional structural regularizers, provides less direct control over latent-space organization; in return, it yields a morphology-design policy pretrained for Morphogene-conditioned generation that can be further optimized during co-design.

\section{Implementation Details}
\label{app:imp_detail}
\subsection{Training Details}
Following standard reinforcement learning practice, we employ distributed trajectory collection across multiple CPU threads to improve training efficiency. All experiments are conducted with five random seeds on a machine equipped with two NVIDIA GeForce RTX 4090 GPUs. The neural networks are implemented in PyTorch 2.0.1, and the morphology--control environments are simulated using MuJoCo 2.1.0~\citep{todorov2012mujoco}.

\subsection{Hyperparameters}
\paragraph{GeCode} 
We implement GeCode in PyTorch on top of BodyGen, retaining its environment, morphology-design space, and policy-optimization settings. 
To construct the Morphogene space, we train the RAE for 100 epochs on randomly generated valid morphologies using Adam with a learning rate of 5e-4 and a batch size of 2{,}048. 
During Morphogene-driven co-design, the morphology encoder remains frozen, while Morphogenes are randomly initialized in the learned latent space and updated after each epoch using existing rollout statistics, without additional environment interactions. 
Detailed hyperparameters are provided in Table~\ref{tab:method_hyperparameters}.
\begin{table}[t]
    \centering
    \caption{Hyperparameters of GeCode adopted in all experiments.}
    \vspace{0.03in}
    \label{tab:method_hyperparameters}
    \setlength{\tabcolsep}{12pt}
    \begin{tabular}{@{\hspace{4pt}}ll@{}}
        \toprule
        \textbf{Hyperparameter} & \textbf{Value} \\
        \midrule
        \rowcolor{gray!15}[2pt][0pt]
        \multicolumn{2}{@{}l@{}}{%
            \textbf{\textsc{Morphogene-space Construction}}%
        } \\
        Space construction epochs  & 100 \\
        RAE batch size & 2{,}048 \\
        RAE Optimizer & Adam \\
        RAE learning rate & 5e-4 \\
        Limb-token dimension & 128 \\
        Morphogene dimension $d_g$ & 32 \\
        Attribute reconstruction loss weight & 1.0 \\
        Limb-presence mask loss weight $\lambda_{\mathrm{mask}}$
            & 1.0 \\
        Morphology-code regularization weight $\lambda_{\mathrm{reg}}$
            & 0.01 \\
        Topology-aware contrastive loss weight $\lambda_{\mathrm{topo}}$ & 8e-3 \\
        Contrastive temperature $\tau_{\text{topo}}$ & 0.1 \\
        Tree-structured metric loss weight $\lambda_{\mathrm{tree}}$ & 0.01 \\
        Tree-structured metric margin $\delta_{\mathrm{tree}}$ & 0.2 \\
        Maximum tree triplets per batch & 64 \\

        \midrule
        \rowcolor{gray!15}[2pt][0pt]
        \multicolumn{2}{@{}l@{}}{%
            \textbf{\textsc{Morphogene-Driven Co-design Optimization}}%
        } \\
          
        Initial Morphogene count $N_G$ & 5 \\
        Similarity threshold $\tau_{\mathrm{sim}}$ & 0.4 \\
        Proximity reward weight $\lambda_{\mathrm{hit}}$ & 1{,}000 \\
        Morphogene update rate $\eta$ & 0.1 \\
        Minimum local hits for retention $H_{\min}$ & 5 \\
        Top-$N_\mathrm{top}$ returns for adaptive sampling ($N_\mathrm{top}$) & 5 \\
        \midrule
        Policy Transformer depth & 3 \\
        Policy Hidden dimension & 64 \\
        Value Transformer depth & 3 \\
        Value Hidden dimension & 64 \\
        Transformer Layer Normalization & Pre-LN \\
        Transformer Activation Function &  SiLU \\

        PPO optimizer & Adam \\
        Policy learning rate & 5e-5 \\
        Value learning rate & 3e-4 \\
        Clip gradient norm & 40.0 \\
        PPO clip $\epsilon$ & 0.2 \\
        PPO batch size & 50{,}000 \\
        PPO minibatch size & 2{,}048 \\
        PPO Iterations Per Batch & 10 \\
        Training Epochs & 1000\\
        Discount factor $\gamma$ & 0.995 \\
        GAE parameter $\lambda_{\text{GAE}}$ & 0.95 \\
        \bottomrule
    \end{tabular}
\vspace{-0.2in}
\end{table}

\paragraph{StackelbergPPO} 
StackelbergPPO extends BodyGen with a Stackelberg optimization framework~\citep{dai2026stackelberg}. We follow the official StackelbergPPO implementation: a Fisher regularization coefficient of $\lambda=\text{5.0}$, up to 20 conjugate-gradient iterations with a relative-error tolerance of $10^{-3}$, six follower-sampling steps per episode, and a gradient-normalization ratio of $\alpha=\text{1.0}$.

\paragraph{BodyGen} 
We follow the official BodyGen implementation~\citep{lu2025bodygen}, using Pre-LN MoSAT blocks with SiLU activations, a hidden dimension of 64, and learning rates of 5e-5 and 3e-4 for the policy and value function, respectively. Additional hyperparameters are provided in Table~\ref{tab:method_hyperparameters}.

\paragraph{Transform2Act} 
We follow the official Transform2Act implementation~\citep{yuan2022transformact}, using three-layer GraphConv policy and value networks with hidden dimensions (64, 64, 64) and learning rates of 5e-5 and 3e-4, respectively. The policy employs a three-layer, 128-unit JSMLP with Tanh activations, while the value function uses a (512, 256) MLP.

\section{Additional Results}
\subsection{Quantitative Results}
\label{app:qua_result}

\paragraph{Quantitative Performance}
Figure~\ref{fig:main_result} presents the learning curves of GeCode and the RL-based baselines, including Transform2Act, BodyGen, and StackelbergPPO, while Table~\ref{tab:main_results} reports the final episode returns as the mean and standard error over five random seeds. GeCode achieves the highest final return in all twelve environments. Relative to the strongest baseline in each task, its improvement ranges from $6.5\%$ on \textsc{Swimmer} to $104.6\%$ on \textsc{Pusher}.

The substantial gains on \textsc{Crawler}, \textsc{Glider-Hard}, and \textsc{Walker-Hard} are particularly notable given their large, highly branched morphology spaces. 
Instead of directly exploring numerous discrete structures, GeCode performs local refinement around individual Morphogene anchors and updates these anchors toward promising regions of the latent space. 
This local-to-global strategy focuses rollouts on structurally coherent neighborhoods while sharing high-performing design information across anchors, thereby improving exploration efficiency as the morphology space expands.

The gains on \textsc{Pusher}, \textsc{Stepper}, and \textsc{TerrainCrosser} further highlight the importance of explicit body--brain coordination. 
By conditioning both morphology generation and control on the same Morphogene, GeCode provides the controller with a compact representation of the structural factors underlying the generated body. 
This shared conditioning enables the controller to adapt its actions to morphology-dependent object contacts, terrain interactions, and stability requirements, while improving credit assignment between body design and control behavior.

In contrast, the smaller gain on \textsc{Swimmer} suggests that explicit body--brain coordination provides a more limited advantage when relatively simple morphologies are already sufficient for effective propulsion. 
Overall, these results show that GeCode derives its advantage from both structured latent-space exploration and explicit morphology--control coordination, with the benefits becoming more pronounced as structural search and control adaptation grow more tightly coupled.

\begin{table*}[ht]
\centering
\setlength{\tabcolsep}{1.8 mm}
\caption{\textbf{Performance comparison across morphology--control co-design environments.} Returns from the final training episode are reported as means $\pm$ standard errors over five random seeds for GeCode, reinforcement-learning baselines, and evolutionary baselines.}
\vspace{0.05in}
\label{tab:main_results}
\resizebox{\linewidth}{!}{
\begin{tabular}{@{}lcccc>{\columncolor{gray!15}}c@{}}
\toprule
\textbf{Task}
& \textbf{NGE} 
& \textbf{Transform2Act} 
& \textbf{BodyGen}
& \textbf{StackelbergPPO}
& \textbf{GeCode} \\

& \citep{wang2018neural}
& \citep{yuan2022transformact}
& \citep{lu2025bodygen}
& \citep{dai2026stackelberg}
& \textbf{(Ours)} \\
\midrule

\textsc{Crawler}
& 1482.5$\pm$525.0
& 5396.3$\pm$177.3
& 8208.6$\pm$374.3
& 9255.3$\pm$888.2
& \textbf{16143.3$\pm$583.0}\\

\textsc{Stepper}
& 870.6$\pm$215.5
& 1671.0$\pm$148.3
& 2087.2$\pm$243.3
& 2538.9$\pm$173.6
& \textbf{3688.6$\pm$271.0} \\

\textsc{Pusher}
& 551.6$\pm$120.7
& 571.8$\pm$78.6
& 1809.1$\pm$162.2
& 3054.4$\pm$354.9
& \textbf{6250.2$\pm$757.9} \\

\textsc{Cheetah}
& 2534.8$\pm$428.7
& 9542.1$\pm$635.0
& 11475.1$\pm$489.5
& 13246.3$\pm$299.9
& \textbf{19888.7$\pm$944.6} \\

\textsc{TerrainCrosser}
& 827.2$\pm$427.2
& 2426.5$\pm$654.1
& 4016.4$\pm$911.6
& 5437.5$\pm$675.6
& \textbf{9670.6$\pm$1176.4} \\

\textsc{Swimmer}
& 384.5$\pm$112.0
& 782.4$\pm$23.6
& 1301.6$\pm$1.3
& 1335.9$\pm$17.2
& \textbf{1422.5$\pm$29.5} \\

\textsc{Glider-Regular}
& 1567.8$\pm$756.7
& 6890.4$\pm$414.8
& 11091.0$\pm$211.8
& 11429.9$\pm$261.1
& \textbf{23317.5$\pm$627.9} \\

\textsc{Glider-Medium}
& 1649.6$\pm$763.6
& 8632.0$\pm$1185.1
& 11016.8$\pm$144.2
& 12652.3$\pm$222.3
& \textbf{22376.4$\pm$809.9} \\

\textsc{Glider-Hard}
& 2081.3$\pm$348.2
& 6218.5$\pm$685.9
& 10973.7$\pm$268.3
& 12110.0$\pm$330.6
& \textbf{19972.8$\pm$332.1} \\

\textsc{Walker-Regular}
& 1402.9$\pm$595.5
& 7960.6$\pm$534.2
& 11555.3$\pm$220.1
& 13216.4$\pm$536.5
& \textbf{24306.1$\pm$1293.0} \\

\textsc{Walker-Medium}
& 2600.4$\pm$481.7
& 7324.4$\pm$906.8
& 13244.4$\pm$342.1
& 11764.6$\pm$418.6
& \textbf{21973.3$\pm$918.9} \\

\textsc{Walker-Hard}
& 1504.6$\pm$553.2
& 4924.0$\pm$690.3
& 12126.0$\pm$825.2
& 13643.4$\pm$597.6
& \textbf{21614.6$\pm$521.8} \\

\bottomrule
\end{tabular}
}
\end{table*}

\paragraph{Comparison with Evolutionary Search}
Table~\ref{tab:main_results} further compares GeCode with NGE~\citep{wang2018neural}, an evolutionary method that optimizes populations of embodied designs through selection and variation. 
GeCode outperforms NGE across all twelve tasks, achieving approximately $3.7\times$ to $17.3\times$ the returns of NGE.
Evolutionary co-design must distribute a finite interaction budget across numerous morphology candidates and evaluate the controller associated with each design. 
As the morphology space becomes larger and more highly branched, this population-level evaluation becomes increasingly expensive, while limited knowledge transfer across structurally distinct candidates hinders the joint identification of effective morphologies and compatible controllers.

In contrast, GeCode organizes co-design around a small set of searchable Morphogene anchors in a compact, structurally regularized latent space. Stochastic policy rollouts refine body--brain designs locally around each anchor, while performance-guided updates move the anchors toward promising regions for global exploration. 
Because all Morphogenes condition the same morphology-design and control policies, knowledge learned from different designs is shared through the policy parameters, while Morphogene updates reuse existing rollout trajectories without additional evaluations.
GeCode therefore replaces costly population-level search with structured latent exploration that directly couples morphology refinement and controller adaptation.

\subsection{Additional Ablation and Analysis}
\begin{table*}[t]
\centering
\setlength{\tabcolsep}{2.5 mm}
\caption{\textbf{RAE training statistics across twelve co-design tasks.}
Values are reported as means $\pm$ standard errors over five random seeds.}
\vspace{0.05in}
\label{tab:RAE}
\resizebox{\linewidth}{!}{
\begin{tabular}{@{}l>{\columncolor{gray!15}}cccccc@{}}
\toprule
\multirow{2}{*}{\textbf{Task}}
& \multirow{2}{*}{$\mathcal{L}_{\mathrm{RAE}}$}
&  \multicolumn{3}{c}{$\mathcal{L}_{\mathrm{rec}}$} 
& \multirow{2}{*}{$\mathcal{L}_{\mathrm{topo}}$}
& \multirow{2}{*}{$\mathcal{L}_{\mathrm{tree}}$}
\\
\cmidrule{3-5}
& 
& $\mathcal{L}_{\mathrm{attr}}$
& $\mathcal{L}_{\mathrm{mask}}$
& $\mathcal{L}_{\mathrm{reg}}$
\\
\midrule
\textsc{Crawler}
& 0.3384 $\pm$ 0.0028
& 0.2928 $\pm$ 0.0025
& 0.0010 $\pm$ 0.0002
& 0.2558 $\pm$ 0.0031
& 5.2215 $\pm$ 0.0627
& 0.0267 $\pm$ 0.0031\\

\textsc{Stepper}
& 0.3336 $\pm$ 0.0058
& 0.2889 $\pm$ 0.0060
& 0.0008 $\pm$ 0.0001
& 0.2509 $\pm$ 0.0085
& 5.1376 $\pm$ 0.0280
& 0.0267 $\pm$ 0.0048 \\

\textsc{Pusher}
& 0.3365 $\pm$ 0.0052
& 0.2911 $\pm$ 0.0053
& 0.0011 $\pm$ 0.0002
& 0.2585 $\pm$ 0.0098
& 5.1803 $\pm$ 0.0310
& 0.0237 $\pm$ 0.0031 \\

\textsc{Cheetah}
& 0.0908 $\pm$ 0.0025
& 0.0496 $\pm$ 0.0024
& 0.0001 $\pm$ 0.0000
& 0.2104 $\pm$ 0.0030
& 4.7858 $\pm$ 0.0152
& 0.0686 $\pm$ 0.0017 \\

\textsc{TerrainCrosser}
& 0.0919 $\pm$ 0.0038
& 0.0496 $\pm$ 0.0033
& 0.0002 $\pm$ 0.0001
& 0.2410 $\pm$ 0.0172
& 4.8627 $\pm$ 0.0554
& 0.0720 $\pm$ 0.0041 \\

\textsc{Swimmer}
& 0.1259 $\pm$ 0.0053
& 0.0797 $\pm$ 0.0052
& 0.0003 $\pm$ 0.0000
& 0.2982 $\pm$ 0.0054
& 5.3088 $\pm$ 0.0431
& 0.0443 $\pm$ 0.0021 \\

\textsc{Glider-Regular}
& 0.0736 $\pm$ 0.0017
& 0.0189 $\pm$ 0.0018
& 0.0000 $\pm$ 0.0000
& 0.1544 $\pm$ 0.0011
& 6.6390 $\pm$ 0.0056
& 0.0000 $\pm$ 0.0000 \\

\textsc{Glider-Medium}
& 0.0873 $\pm$ 0.0138
& 0.0361 $\pm$ 0.0128
& 0.0001 $\pm$ 0.0000
& 0.2098 $\pm$ 0.0174
& 6.1070 $\pm$ 0.0800
& 0.0229 $\pm$ 0.0139 \\

\textsc{Glider-Hard}
& 0.0870 $\pm$ 0.0079
& 0.0387 $\pm$ 0.0072
& 0.0001 $\pm$ 0.0000
& 0.2459 $\pm$ 0.0252
& 5.6611 $\pm$ 0.0474
& 0.0357 $\pm$ 0.0030 \\

\textsc{Walker-Regular}
& 0.0783 $\pm$ 0.0029
& 0.0285 $\pm$ 0.0029
& 0.0001 $\pm$ 0.0000
& 0.1782 $\pm$ 0.0038
& 5.9831 $\pm$ 0.0055
& 0.0087 $\pm$ 0.0011 \\

\textsc{Walker-Medium}
& 0.1166 $\pm$ 0.0030
& 0.0741 $\pm$ 0.0026
& 0.0003 $\pm$ 0.0000
& 0.2308 $\pm$ 0.0111
& 4.8997 $\pm$ 0.0504
& 0.0750 $\pm$ 0.0041 \\

\textsc{Walker-Hard}
& 0.1793 $\pm$ 0.0048
& 0.1380 $\pm$ 0.0045
& 0.0007 $\pm$ 0.0003
& 0.2334 $\pm$ 0.0053
& 4.7300 $\pm$ 0.0191
& 0.0423 $\pm$ 0.0035 \\

\bottomrule
\end{tabular}
}
\end{table*}

\paragraph{Analysis of RAE Training}
Table~\ref{tab:RAE} reports the task-specific RAE training losses. The consistently small standard errors of the overall objective $\mathcal{L}_{\mathrm{RAE}}$ indicate stable optimization across all twelve environments. The near-zero limb-presence loss $\mathcal{L}_{\mathrm{mask}}$ shows that the decoder reliably reconstructs variable limb configurations, which is essential for representing morphologies with different topologies. The attribute-reconstruction loss $\mathcal{L}_{\mathrm{attr}}$ is generally higher for the three-dimensional tasks, including \textsc{Crawler}, \textsc{Stepper}, and \textsc{Pusher}, reflecting the greater complexity of their geometric representations. 
Across the \textsc{Glider} and \textsc{Walker} task groups, $\mathcal{L}_{\mathrm{attr}}$ generally increases as the branching constraints are relaxed, consistent with the greater morphological diversity admitted by the more challenging variants.

The topology-aware contrastive loss $\mathcal{L}_{\mathrm{topo}}$ and tree-metric loss $\mathcal{L}_{\mathrm{tree}}$ impose complementary structure on the latent space. 
Specifically, $\mathcal{L}_{\mathrm{topo}}$ organizes morphology codes into coherent local neighborhoods according to topology, preventing structurally distinct morphologies from becoming arbitrarily entangled. In contrast, $\mathcal{L}_{\mathrm{tree}}$ preserves graded structural relationships by ordering latent distances according to tree-edit similarity, such that nearby codes correspond to progressively related morphology structures. 
The consistently small standard errors indicate stable optimization of both objectives across environments. Moreover, the near-zero $\mathcal{L}_{\mathrm{tree}}$ on weakly branching tasks such as \textsc{Glider-Regular} and its larger values in more structurally diverse tasks reflect the increasing difficulty of preserving global structural order as branching complexity grows.
Together, these losses balance topology-aware local clustering with globally meaningful distance organization, enabling Morphogene updates to traverse the latent space through coherent structural variations.

\begin{figure*}[p]
  \centering
  \includegraphics[width=\linewidth]{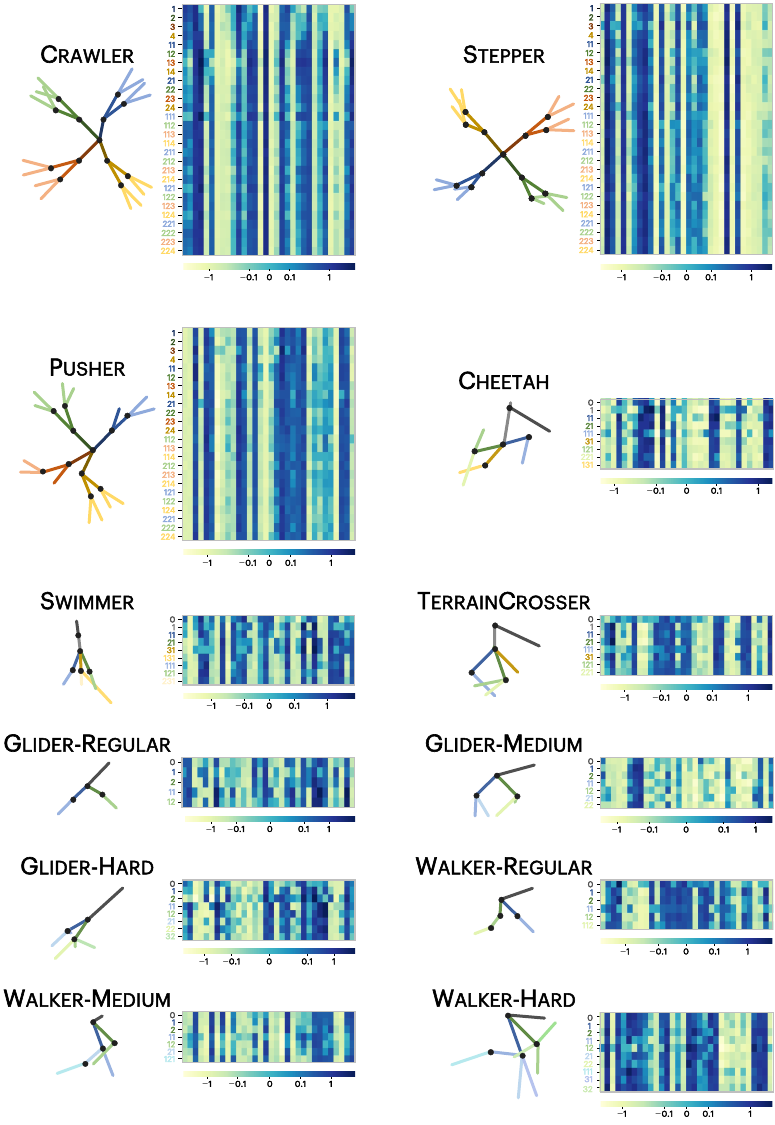}
  \vspace{-0.3in}
  \caption{\textbf{Limb-specific Morphogene conditioning with AdaConcat.} 
  Representative morphologies are paired with heatmaps of their limb-level conditions. Distinct patterns across rows show how AdaConcat adapts a shared Morphogene to individual limbs.}
  \label{fig:adaconcat}
  \vspace{-0.1in}
\end{figure*}

\begin{figure*}[tb]
  \centering
  \includegraphics[width=\linewidth]{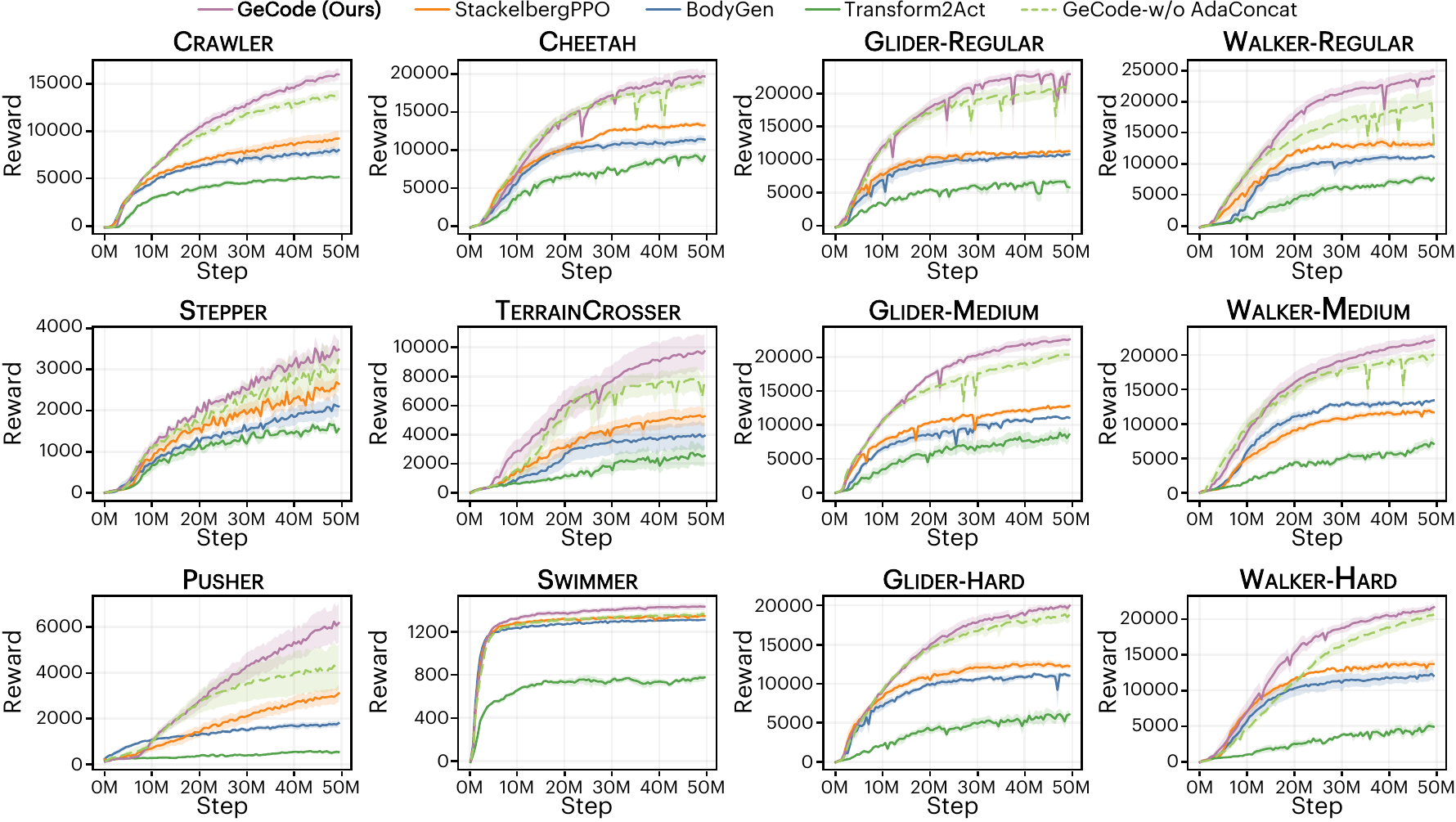}
  \vspace{-0.2in}
  \caption{\textbf{Ablation of AdaConcat in morphology--control co-design.} Learning curves compare GeCode, its direct-concatenation variant (i.e., GeCode w/o AdaConcat), and reference baselines across twelve tasks. Shaded regions indicate the standard error over five random seeds.}
  \label{fig:adaconcat_curve}
  \vspace{-0.1in}
\end{figure*}

\begin{table*}[t]
\centering
\setlength{\tabcolsep}{2 mm}
\caption{\textbf{Quantifying limb-specific Morphogene conditioning.} Limb Diversity, Relative Modulation, and Between-Limb Energy Ratio measure differences among limb-level conditions under AdaConcat and direct concatenation across twelve tasks. Values are reported as means $\pm$ standard errors over five random seeds.}
\vspace{0.07in}
\label{tab:adaconcat_modulation}
\resizebox{\textwidth}{!}{%
\begin{tabular}{@{}lcc|cc|cc@{}}
\toprule
& \multicolumn{2}{c}{\textbf{Limb Diversity}}
& \multicolumn{2}{c}{\textbf{Relative Modulation}}
& \multicolumn{2}{c}{\textbf{Between-Limb Energy Ratio}} \\
\cmidrule(lr){2-3}\cmidrule(lr){4-5}\cmidrule(l){6-7}
Task & \textbf{AdaConcat} & Direct Concat & \textbf{AdaConcat} & Direct Concat & \textbf{AdaConcat} &  Direct Concat \\
\midrule
\textsc{Crawler}       
& \textbf{0.658 $\pm$ 0.053}
& 0.000 
& \textbf{0.482 $\pm$ 0.041}
& 0.000 
& \textbf{0.181 $\pm$ 0.020}
& 0.000 \\
\textsc{Stepper}      
& \textbf{0.746 $\pm$ 0.126} 
& 0.000 
& \textbf{0.542 $\pm$ 0.096} 
& 0.000 
& \textbf{0.203 $\pm$ 0.049}
& 0.000 \\
\textsc{Pusher}         
& \textbf{0.643 $\pm$ 0.047} 
& 0.000 
& \textbf{0.480 $\pm$ 0.036} 
& 0.000 
& \textbf{0.165 $\pm$ 0.017} 
& 0.000 \\
\textsc{Cheetah}    
& \textbf{1.257 $\pm$ 0.081}
& 0.000
& \textbf{0.841 $\pm$ 0.050}
& 0.000 
& \textbf{0.346 $\pm$ 0.034}
& 0.000 \\
\textsc{Terraincrosser}
& \textbf{1.328 $\pm$ 0.119}
& 0.000 
& \textbf{0.894 $\pm$ 0.076}
& 0.000 
& \textbf{0.365 $\pm$ 0.025} 
& 0.000 \\
\textsc{Swimmer}      
& \textbf{1.169 $\pm$ 0.165}
& 0.000 
& \textbf{0.772 $\pm$ 0.110}
& 0.000 
& \textbf{0.334 $\pm$ 0.035} 
& 0.000 \\
\textsc{Glider-Regular} 
& \textbf{2.288 $\pm$ 0.537}
& 0.000 
& \textbf{1.522 $\pm$ 0.309}
& 0.000 
& \textbf{0.410 $\pm$ 0.068} 
& 0.000 \\
\textsc{Glider-Medium}
& \textbf{1.328 $\pm$ 0.191} 
& 0.000 
& \textbf{0.876 $\pm$ 0.106}
& 0.000 
& \textbf{0.350 $\pm$ 0.061}
& 0.000 \\
\textsc{Glider-Hard}    
& \textbf{1.319 $\pm$ 0.362}
& 0.000 
& \textbf{0.862 $\pm$ 0.229}
& 0.000 
& \textbf{0.336 $\pm$ 0.071} 
& 0.000 \\
\textsc{Walker-Regular} 
& \textbf{1.453 $\pm$ 0.140}
& 0.000 
& \textbf{0.968 $\pm$ 0.070}
& 0.000
& \textbf{0.384 $\pm$ 0.037}
& 0.000 \\
\textsc{Walker-Medium}  
& \textbf{1.031 $\pm$ 0.136} 
& 0.000 
& \textbf{0.714 $\pm$ 0.112}
& 0.000 
& \textbf{0.290 $\pm$ 0.039}
& 0.000 \\
\textsc{Walker-Hard}    
& \textbf{1.275 $\pm$ 0.090} 
& 0.000 
& \textbf{0.865 $\pm$ 0.076}
& 0.000 
& \textbf{0.369 $\pm$ 0.016}
& 0.000 \\
\bottomrule
\end{tabular}
}
\end{table*}

\paragraph{Analysis of Limb-Specific Conditioning in AdaConcat}
Figure~\ref{fig:adaconcat} shows that AdaConcat produces distinct limb-level conditions from the same Morphogene while preserving feature patterns shared across the body. By contrast, direct concatenation provides every limb with an identical Morphogene, requiring the policies to infer limb-specific distinctions from other inputs. 
Complementing the performance comparison in Table~\ref{tab:adaconcat}, the learning curves in Figure~\ref{fig:adaconcat_curve} show that GeCode generally outperforms this direct-concatenation variant, with pronounced gains on \textsc{Pusher} and \textsc{Walker-Regular}. 
These results suggest that adapting the shared Morphogene to each limb helps the policies account for differing roles, such as object contact and locomotor support, during morphology design and control.

To quantify the limb specificity of AdaConcat, we use three complementary measures.
Let $\mathcal{G}_{l_i}$ denote the limb-specific condition produced by AdaConcat from $\mathcal{G}$ for limb $l_i$.
Limb Diversity is the mean pairwise distance between limb conditions, normalized by the magnitude of $\mathcal{G}$:
\begin{equation*}
D_{\mathrm{limb}} = 
\frac{2}{N_L(N_L-1)}
\sum_{i<j}
\frac{\|\mathcal{G}_{l_i}-\mathcal{G}_{l_j}\|_2}
{\|\mathcal{G}\|_2}.
\end{equation*}
Relative Modulation measures the average deviation of the limb-specific conditions from the shared Morphogene, normalized by its magnitude:
\begin{equation*}
D_{\mathrm{mod}} = 
\frac{1}{N_L}
\sum_{i=1}^{N_L}
\frac{\|\mathcal{G}_{l_i}-\mathcal{G}\|_2}
{\|\mathcal{G}\|_2}.
\end{equation*}
Between-Limb Energy Ratio measures the proportion of total condition energy associated with variation across limbs:
\begin{equation*}
R_{\mathrm{between}} = 
\frac{\sum_{i=1}^{N_L}
\|\mathcal{G}_{l_i}-\boldsymbol{\mu}_{\mathrm{limb}}\|_2^2}
{\sum_{i=1}^{N_L}\|\mathcal{G}_{l_i}\|_2^2},
\qquad
\boldsymbol{\mu}_{\mathrm{limb}} = 
\frac{1}{N_L}\sum_{i=1}^{N_L}\mathcal{G}_{l_i}.
\end{equation*}

Table~\ref{tab:adaconcat_modulation} shows that AdaConcat consistently produces distinct limb-specific conditions from a shared Morphogene. 
Nonzero Relative Modulation indicates that these conditions differ from the shared Morphogene, while nonzero Limb Diversity and Between-Limb Energy Ratio show that the transformation varies across limbs. Between-limb variation accounts for $16.5\%$–$41.0\%$ of total condition energy, indicating substantial limb specificity while retaining a shared component. 
In contrast, all three measures are zero under direct concatenation, which provides every limb with the same unmodified Morphogene.

\begin{table*}[t]
\centering
\caption{\textbf{Environment interactions required to reach baseline performance thresholds.}
For each task, the two thresholds are the highest returns attained during training by BodyGen~\citep{lu2025bodygen} and StackelbergPPO~\citep{dai2026stackelberg}, respectively.
Entries report the required interactions in millions (M); $\infty$ indicates that a method did not reach the threshold within the 50M-interaction budget. The lowest interaction count in each row is shown in \textbf{bold}.}
\vspace{0.07in}
\label{tab:sample_efficiency}
\resizebox{\textwidth}{!}{
\begin{tabular}{@{}l||c||>{\columncolor{gray!15}}cccccc@{}}
\toprule
\textbf{Task} & \shortstack[c]{\textbf{Threshold}\\\textbf{(BodyGen)}}
& \textbf{GeCode}
& \textbf{StackelbergPPO}
& \textbf{BodyGen}
& \textbf{Transform2Act}
& \textbf{NGE}
& \textbf{ESS}
\\
\midrule

\textsc{Crawler}
& 8208.57
& \textbf{14.35 M}
& 31.10 M
& 49.45 M
& $\infty$
& $\infty$
& $\infty$
\\

\textsc{Stepper}
& 2087.16
& \textbf{20.90 M}
& 31.20 M
& 45.80 M
& $\infty$
& $\infty$
& $\infty$
\\

\textsc{Pusher}
& 1809.06
& \textbf{14.20 M}
& 24.20 M
& 48.20 M
& $\infty$
& $\infty$
& $\infty$
\\

\textsc{Cheetah}
& 11475.06
& \textbf{15.05 M}
& 25.15 M
& 45.40 M
& $\infty$
& $\infty$
& $\infty$
\\

\textsc{TerrainCrosser}
& 4016.37
& \textbf{14.05 M}
& 24.75 M
& 50.00 M
& $\infty$
& $\infty$
& $\infty$
\\

\textsc{Swimmer}
& 1301.57
& \textbf{8.05 M}
& 14.70 M
& 36.75 M
& $\infty$
& $\infty$
& $\infty$
\\

\textsc{Glider-Regular}
& 11091.03
& \textbf{8.65 M}
& 28.00 M
& 49.85 M
& $\infty$
& $\infty$
& $\infty$
\\

\textsc{Glider-Medium}
& 11016.82
& \textbf{10.55 M}
& 26.10 M
& 44.75 M
& $\infty$
& $\infty$
& $\infty$
\\

\textsc{Glider-Hard}
& 10973.74
& \textbf{12.65 M}
& 19.70 M
& 40.30 M
& $\infty$
& $\infty$
& $\infty$
\\

\textsc{Walker-Regular}
& 11555.34
& \textbf{12.90 M}
& 18.80 M
& 50.00 M
& $\infty$
& $\infty$
& $\infty$
\\

\textsc{Walker-Medium}
& 13244.43
& \textbf{15.35 M}
& $\infty$
& 46.25 M
& $\infty$
& $\infty$
& $\infty$
\\

\textsc{Walker-Hard}
& 12125.98
& \textbf{15.45 M}
& 22.25 M
& 46.55 M
& $\infty$
& $\infty$
& $\infty$
\\

\midrule
\textbf{Task} & \shortstack[c]{\textbf{Threshold}\\\textbf{(StackelbergPPO)}}
& \textbf{GeCode}
& \textbf{StackelbergPPO}
& \textbf{BodyGen}
& \textbf{Transform2Act}
& \textbf{NGE}
& \textbf{ESS}
\\
\midrule
\textsc{Crawler}
& 9255.29
& \textbf{16.30 M}
& 46.30 M
& $\infty$
& $\infty$
& $\infty$
& $\infty$
\\

\textsc{Stepper}
& 2538.91
& \textbf{26.65 M}
& 42.75 M
& $\infty$
& $\infty$
& $\infty$
& $\infty$
\\

\textsc{Pusher}
& 3054.39
& \textbf{21.40 M}
& 48.85 M
& $\infty$
& $\infty$
& $\infty$
& $\infty$
\\

\textsc{Cheetah}
& 13246.30
& \textbf{17.55 M}
& 37.25 M
& $\infty$
& $\infty$
& $\infty$
& $\infty$
\\

\textsc{TerrainCrosser}
& 5437.52
& \textbf{18.35 M}
& 50.00 M
& $\infty$
& $\infty$
& $\infty$
& $\infty$
\\

\textsc{Swimmer}
& 1335.90
& \textbf{11.20 M}
& 37.05 M
& $\infty$
& $\infty$
& $\infty$
& $\infty$
\\

\textsc{Glider-Regular}
& 11429.94
& \textbf{9.20 M}
& 30.35 M
& $\infty$
& $\infty$
& $\infty$
& $\infty$
\\

\textsc{Glider-Medium}
& 12652.30
& \textbf{12.80 M}
& 44.35 M
& $\infty$
& $\infty$
& $\infty$
& $\infty$
\\

\textsc{Glider-Hard}
& 12110.01
& \textbf{14.75 M}
& 29.60 M
& $\infty$
& $\infty$
& $\infty$
& $\infty$
\\

\textsc{Walker-Regular}
& 13216.44
& \textbf{14.70 M}
& 27.85 M
& $\infty$
& $\infty$
& $\infty$
& $\infty$
\\

\textsc{Walker-Medium}
& 11764.57
& \textbf{13.65 M}
& 43.05 M
& 24.05 M
& $\infty$
& $\infty$
& $\infty$
\\

\textsc{Walker-Hard}
& 13643.36
& \textbf{17.10 M}
& 39.30 M
& $\infty$
& $\infty$
& $\infty$
& $\infty$
\\

\bottomrule
\end{tabular}
}
\end{table*}

\subsection{Sample and Training Efficiency}
\label{sec:app_efficient}
\paragraph{Sample Efficiency}
We measure sample efficiency by the number of environment interactions required to reach task-specific thresholds, defined by the highest returns attained during training by BodyGen and StackelbergPPO.
As shown in Table~\ref{tab:sample_efficiency}, GeCode reaches both thresholds with the fewest interactions in all twelve tasks. Summed across tasks, GeCode requires 48.06\% and 57.69\% fewer interactions than the fastest task-specific baseline to reach the BodyGen and StackelbergPPO thresholds, respectively.
For example, it reaches the BodyGen threshold on \textsc{Glider-Regular} in 8.65M interactions, compared with 28.00M for the fastest baseline. 
GeCode also reaches every StackelbergPPO threshold within the 50M-interaction budget, whereas BodyGen reaches only one.

This efficiency arises from the complementary roles of local rollouts and global Morphogene updates.
Rollouts explore related body designs around each Morphogene, providing more consistent experience for the conditioned controller. 
Performance-guided updates and adaptive sampling subsequently direct interactions toward promising regions of the Morphogene space.
Because these operations reuse policy-training trajectories, they improve the focus of exploration without additional environment interactions.

\begin{table}[t]
\centering
\caption{\textbf{Training-time computational cost across morphology design spaces.}
Wall-clock time is reported for offline RAE training, morphology--control co-design, and their combined cost in GeCode, together with the training time of StackelbergPPO. The maximum number of limbs indicates the capacity of each task-specific design space. 
Values are reported as means $\pm$ standard errors over five random seeds.}
\vspace{0.05in}
\label{tab:app_wall_clock}
\resizebox{\linewidth}{!}{%
\begin{tabular}{@{}lccccc@{}}
\toprule
\textbf{Task}
& \textbf{Space Size (max)}
& \multicolumn{3}{c}{\textbf{GeCode}}
& \textbf{StackelbergPPO} \\
\cmidrule(l){3-5}
& & Offline RAE Training & Co-design Optimization & Total Time & \\
\midrule
\textsc{Crawler}
& 29
& 0.997 $\pm$ 0.017 & 30.755 $\pm$ 3.390 & 31.752 $\pm$ 3.389
& \textbf{27.924 $\pm$ 1.617} \\

\textsc{Stepper}
& 29
& 1.027 $\pm$ 0.020 & 29.560 $\pm$ 3.583 & \textbf{30.587 $\pm$ 3.566}
& 30.969 $\pm$ 1.346 \\

\textsc{Pusher}
& 29
& 1.028 $\pm$ 0.022 & 22.754 $\pm$ 4.045 & \textbf{23.782 $\pm$ 4.031}
& 25.447 $\pm$ 1.811 \\

\textsc{Cheetah}
& 14
& 0.767 $\pm$ 0.003 & 21.384 $\pm$ 0.676 & 22.151 $\pm$ 0.679
& \textbf{19.959 $\pm$ 0.205} \\

\textsc{TerrainCrosser}
& 14
& 0.883 $\pm$ 0.023 & 24.279 $\pm$ 1.040 & 25.161 $\pm$ 1.033
& \textbf{24.020 $\pm$ 1.441} \\

\textsc{Swimmer}
& 14
& 0.839 $\pm$ 0.017 & 17.948 $\pm$ 1.214 & \textbf{18.787 $\pm$ 1.220}
& 20.913 $\pm$ 0.748 \\

\textsc{Glider-Regular}
& 5
& 0.877 $\pm$ 0.024 & 19.320 $\pm$ 0.809 & 20.197 $\pm$ 0.789
& \textbf{19.319 $\pm$ 0.379} \\

\textsc{Glider-Medium}
& 7
& 0.861 $\pm$ 0.025 & 19.316 $\pm$ 0.212 & \textbf{20.178 $\pm$ 0.211}
& 20.494 $\pm$ 0.951 \\

\textsc{Glider-Hard}
& 9
& 0.839 $\pm$ 0.023 & 19.728 $\pm$ 0.805 & \textbf{20.567 $\pm$ 0.787}
& 21.072 $\pm$ 0.939 \\

\textsc{Walker-Regular}
& 7
& 0.878 $\pm$ 0.020 & 18.338 $\pm$ 0.692 & \textbf{19.216 $\pm$ 0.685}
& 22.691 $\pm$ 0.417 \\

\textsc{Walker-Medium}
& 15
& 0.823 $\pm$ 0.023 & 18.960 $\pm$ 0.424 & \textbf{19.783 $\pm$ 0.405}
& 24.666 $\pm$ 0.899 \\

\textsc{Walker-Hard}
& 27
& 1.014 $\pm$ 0.016 & 22.248 $\pm$ 2.395 & \textbf{23.263 $\pm$ 2.388}
& 23.574 $\pm$ 0.476 \\
\midrule
\multicolumn{4}{c}{\textsc{\textbf{Average}}} & \textbf{22.952 $\pm$ 0.760} &  23.421 $\pm$ 0.514 \\ 
\bottomrule
\end{tabular}%
}
\end{table}

\begin{table*}[t]
\centering
\caption{\textbf{Evaluation performance and inference-time computational cost of GeCode.}
Evaluation returns, parameter counts, inference latency, total FLOPs, and per-limb FLOPs are reported across twelve tasks. 
For each random seed, the evaluation return is averaged over five episodes, while latency and FLOPs are measured per control decision. Results are reported as means $\pm$ standard errors over five random seeds.}
\vspace{0.05in}
\label{tab:inference_cost}
\resizebox{\textwidth}{!}{%
\begin{tabular}{@{}lccccc@{}}
\toprule
\textbf{Task}
& \textbf{Eval. Return}
& \textbf{Params. (M)}
& \textbf{Latency (ms)} $\downarrow$
& \textbf{FLOPs (M)} $\downarrow$
& \textbf{FLOPs (Per Limb) (M)} $\downarrow$ \\
\midrule
\textsc{Crawler}
& 15899.313 $\pm$ 619.609 & 0.5522
& 16.734 $\pm$ 0.759 & 9.354 & 0.323 \\

\textsc{Stepper}
& 3336.782 $\pm$ 335.130 & 0.5522
& 31.431 $\pm$ 8.901 & 7.617 & 0.319 \\

\textsc{Pusher}
& 6019.497 $\pm$ 688.621 & 0.5524
& 13.951 $\pm$ 0.970 & 8.682 & 0.321 \\

\textsc{Cheetah}
& 20130.416 $\pm$ 1012.861 & 0.5517
& 1.511 $\pm$ 0.766 & 2.195 & 0.305 \\

\textsc{TerrainCrosser}
& 9760.178 $\pm$ 1195.632 & 0.5518
& 12.976 $\pm$ 11.144 & 2.385 & 0.305 \\

\textsc{Swimmer}
& 1438.432 $\pm$ 25.512 & 0.5517
& 1.743 $\pm$ 1.113 & 2.319 & 0.305 \\

\textsc{Glider-Regular}
& 23770.548 $\pm$ 668.018 & 0.5515
& 0.635 $\pm$ 0.049 & 1.514 & 0.303 \\

\textsc{Glider-Medium}
& 22664.985 $\pm$ 821.074 & 0.5515
& 0.607 $\pm$ 0.032 & 1.945 & 0.304 \\

\textsc{Glider-Hard}
& 20096.701 $\pm$ 302.635 & 0.5515
& 0.616 $\pm$ 0.002 & 2.441 & 0.305 \\

\textsc{Walker-Regular}
& 24500.695 $\pm$ 1315.028 & 0.5517
& 0.614 $\pm$ 0.039 & 2.070 & 0.304 \\

\textsc{Walker-Medium}
& 22144.726 $\pm$ 950.681 & 0.5517
& 14.692 $\pm$ 12.166 & 2.824 & 0.306 \\

\textsc{Walker-Hard}
& 21732.894 $\pm$ 543.298 & 0.5517
& 45.465 $\pm$ 12.229 & 3.072 & 0.307 \\
\bottomrule
\end{tabular}%
}
\end{table*}

\paragraph{Training Efficiency}
GeCode maintains competitive training efficiency, with wall-clock costs comparable to—and often lower than—those of StackelbergPPO across the evaluated tasks.
As shown in Table~\ref{tab:app_wall_clock}, offline RAE training requires only between $0.77$ and $1.03$ hours of wall-clock time, accounting for less than $4\%$ of GeCode’s overall training time and thus introducing minimal computational overhead.
During co-design, Morphogene updates reuse trajectories collected for policy optimization and involve only morphology encoding and lightweight latent-space operations, requiring neither additional environment interactions nor nested policy-optimization procedures.

Table~\ref{tab:app_wall_clock} shows that GeCode incurs moderately higher wall-clock costs on three-dimensional tasks such as \textsc{Crawler}, while demonstrating a clear efficiency advantage on two-dimensional planar tasks, particularly \textsc{Glider} and \textsc{Walker}.
The additional cost on three-dimensional tasks arises primarily from the greater complexity of the morphologies discovered by GeCode, rather than from the Morphogene update procedure itself. 
By exploring broader regions of the design space, GeCode identifies higher-performing and structurally richer agents with more limbs than StackelbergPPO. These larger morphologies increase simulation cost and produce longer limb-token sequences for Transformer processing, thereby extending policy-optimization time. 
\textbf{The increased wall-clock cost therefore reflects GeCode’s ability to explore more complex morphology regions, rather than substantial overhead from its latent-space search mechanism.}

On two-dimensional planar tasks such as \textsc{Walker}, where the generated morphologies are comparable in scale to those of StackelbergPPO, GeCode incurs lower computational cost. 
This advantage reflects the computational efficiency of Morphogene-space optimization.
Unlike StackelbergPPO, which incorporates follower updates and second-order optimization, GeCode performs lightweight, performance-guided latent updates using existing rollout statistics. Its optimization overhead therefore remains small, with the overall computational cost determined primarily by morphology complexity rather than the Morphogene update procedure.

\paragraph{Inference Efficiency}
Table~\ref{tab:inference_cost} demonstrates that GeCode remains computationally lightweight during evaluation. Its parameter count is nearly constant across tasks at approximately $0.552$M, reflecting the use of shared morphology-design and control networks rather than morphology-specific models. 
Total computation ranges from $1.514$M to $9.354$M FLOPs per control decision, primarily reflecting differences in the number of limbs processed. 
Notably, the per-limb cost remains within a narrow range of $0.303$M--$0.323$M FLOPs, indicating that \textbf{inference scales predictably with morphology size}. 
GeCode therefore supports structurally diverse agents without introducing design-specific networks or substantial inference overhead.

\subsection{Visualization}
\label{app:visual}
\begin{figure*}[p]
  \centering
  \includegraphics[width=0.93\linewidth]{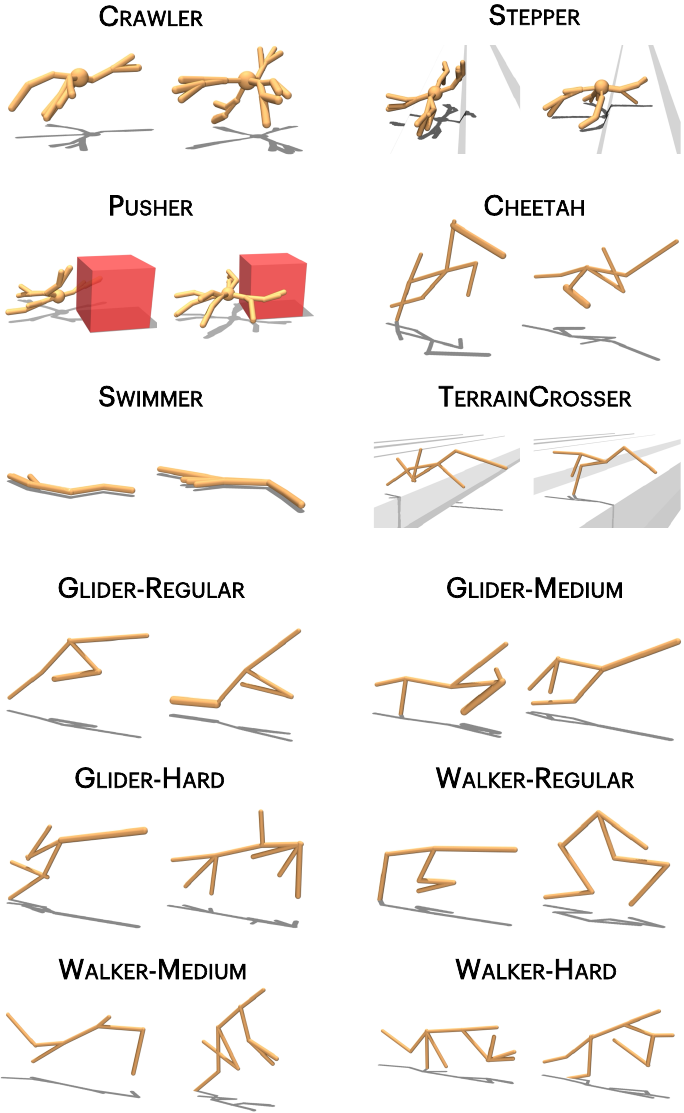}
  \vspace{-0.05in}
  \caption{\textbf{Representative morphologies discovered by GeCode across the co-design environments.}
  The generated agents exhibit task-specific variations in topology, limb configuration, and body proportions, demonstrating the structural diversity enabled by Morphogene-driven co-design.}
  \label{fig:visual_morph}
\end{figure*}

\paragraph{Morphologies Discovered by GeCode}
Figure~\ref{fig:visual_morph} presents representative morphologies discovered by GeCode across the evaluated environments. 
The generated agents exhibit substantial diversity in topology, limb arrangement, and body proportions, ranging from compact locomotor structures to highly branched designs for complex-terrain traversal and object manipulation. 
Rather than remaining confined to minor variations of the initial morphology, Morphogene-driven exploration enables GeCode to reach structurally distinct yet task-appropriate regions of the design space. 
The diversity and validity of these agents demonstrate that GeCode provides broad morphology-space coverage while preserving the realizability and functional relevance of the discovered designs.

\begin{figure*}[p]
  \centering
  \includegraphics[width=\linewidth]{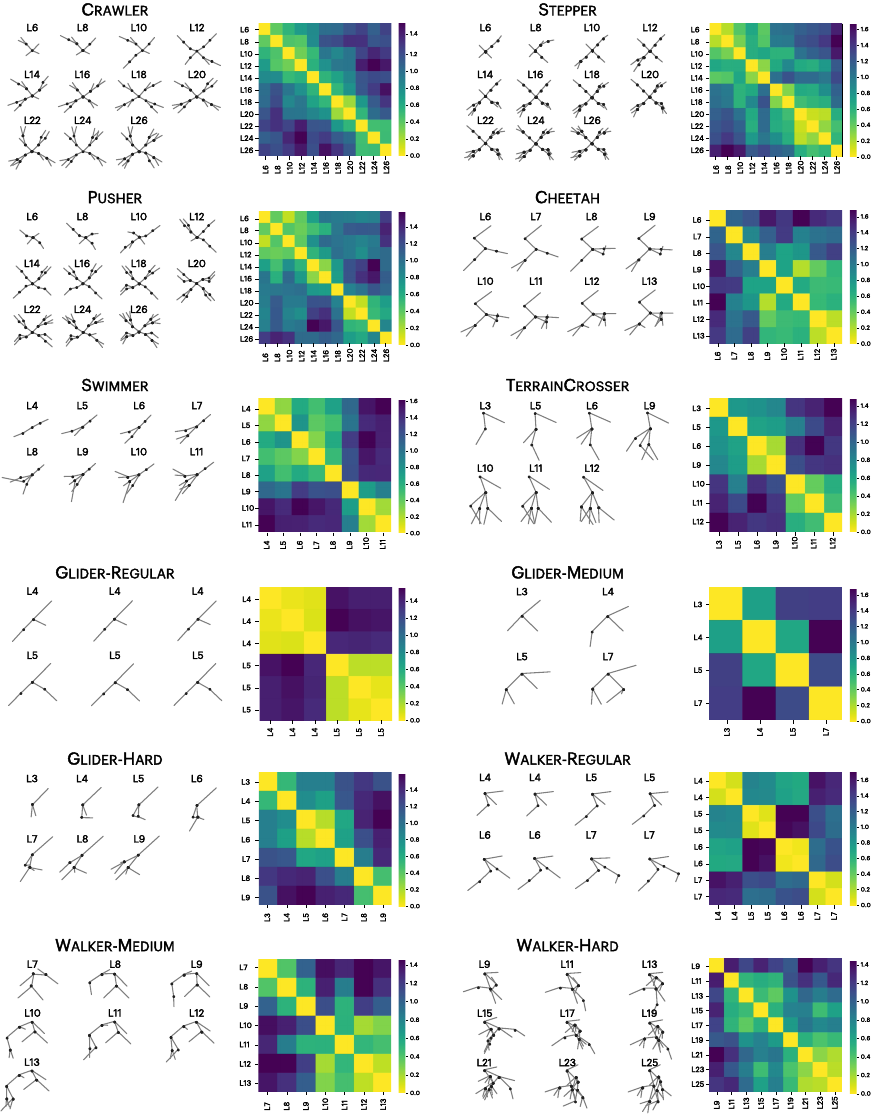}
  \vspace{-0.2in}
  \caption{\textbf{Structural organization of morphology codes in the Morphogene space.}
  Representative morphologies are shown together with their pairwise cosine-distance matrices, where $L_n$ denotes the number of limbs. 
  Lighter colors indicate greater latent similarity, whereas darker colors indicate larger distances. 
  The block and gradual patterns demonstrate that morphologies with similar complexity, topology, and limb configurations are organized into nearby regions of the latent space.}
  \label{fig:morph_similarity}
\end{figure*}

\paragraph{Visualization of Structural Organization in the Morphogene Space}
Figure~\ref{fig:morph_similarity} examines whether distances between morphology codes reflect meaningful structural relationships. The pairwise distance matrices exhibit clear block and graded patterns aligned with morphology complexity, topology, and limb configuration. Morphologies with similar limb counts and branching structures form compact latent neighborhoods, whereas structurally distinct designs are separated by larger distances. This organization is particularly evident in the \textsc{Glider} variants, where morphologies with the same limb count form low-distance blocks. In broader design spaces, such as \textsc{Crawler}, \textsc{Stepper}, and \textsc{Walker-Hard}, latent distances generally increase as structural complexity and branching patterns diverge. Variations within the same limb-count group further show that the representation captures finer geometric and connectivity differences beyond body count.
These results demonstrate that the learned latent space places structurally similar morphologies close together while separating increasingly different designs. 
This organization supports Morphogene-driven co-design by enabling coherent local exploration around each Morphogene and structured global exploration through performance-guided updates.

\begin{figure*}[p]
  \centering
  \includegraphics[width=0.95\linewidth]{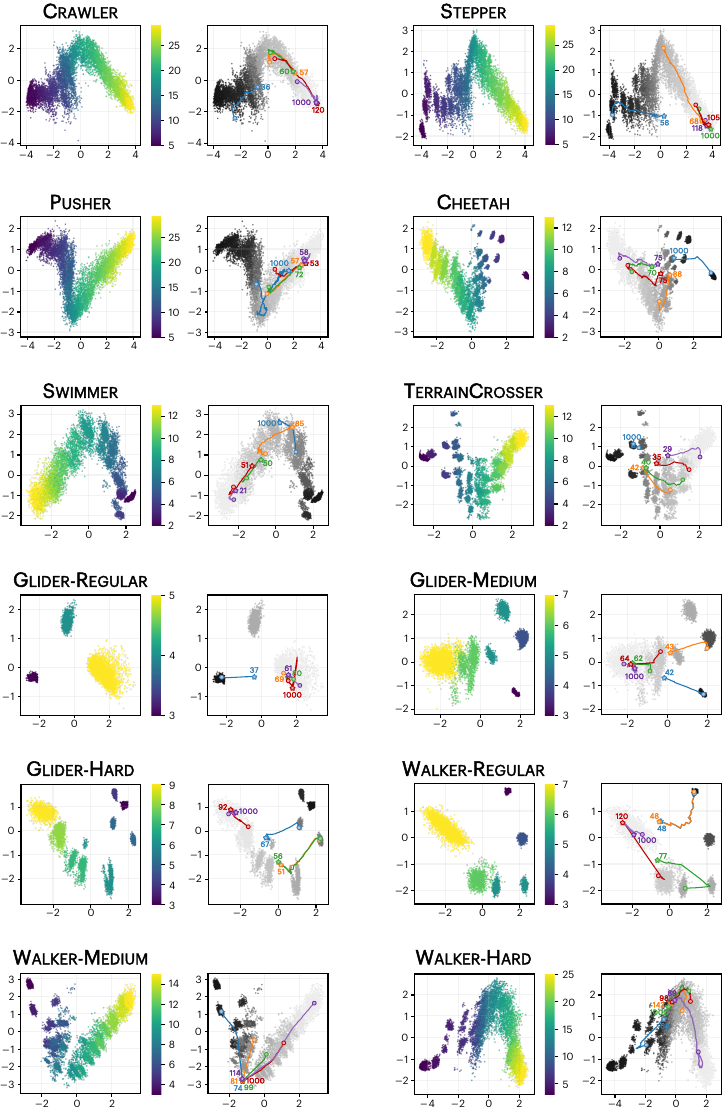}
  \vspace{-0.1in}
  \caption{\textbf{Morphogene update trajectories in task-specific latent spaces.}
  For each environment, the left panel shows a PCA projection of valid morphology codes, while the right panel overlays Morphogene trajectories during co-design. These trajectories illustrate how performance-guided updates explore complementary regions of the structured latent space. \mycircle{} and \mystar{} mark the initial and final positions, respectively. Numerical labels indicate the epoch at which each Morphogene is eliminated or, for retained Morphogenes, the final epoch of the 1,000-epoch co-design process.}
  \label{fig:gene_trajectory}
\end{figure*}

\paragraph{Visualization of Morphogene Update Trajectories}
Figure~\ref{fig:gene_trajectory} shows that valid morphology codes form distinct clusters associated with structural complexity, indicating that the learned latent spaces preserve meaningful morphological relationships. 
During co-design, Morphogenes follow task-dependent trajectories through this structured support rather than moving as unconstrained latent perturbations. In broad design spaces, such as \textsc{Crawler}, \textsc{Cheetah}, and \textsc{Walker-Hard}, different Morphogenes traverse complementary regions and reach morphology families with varying structural complexity. Progressive elimination of underperforming Morphogenes then concentrates the rollout budget on promising regions while retaining diversity among the surviving trajectories.

These trajectories reveal how GeCode integrates local and global exploration. Stochastic rollouts refine morphology variations around each Morphogene, whereas performance-guided updates move the Morphogenes across broader regions of the latent space. Consequently, a small Morphogene set can efficiently cover diverse yet structurally coherent morphology families without exhaustively evaluating independent designs.

\section{Discussion}

\subsection{Comparison with UNIMAL}
\paragraph{Morphology Design Spaces} 
UNIMAL~\citep{gupta2021embodied} provides a well-structured design space for articulated agents, with constrained limb addition, deletion, and continuous attribute mutations that support diverse yet physically valid morphologies. Its explicit mutation operators offer a systematic basis for studying how body structure affects locomotion performance.
The BodyGen-based benchmark adopted in this work supports comparable design operations but evaluates co-design across twelve task-specific environments spanning 2D and 3D locomotion, uneven-terrain traversal, and object manipulation.
Each environment specifies its own initial morphology, task objective, and structural constraints. Together, these settings admit a wider variety of viable designs—from compact planar agents to larger, highly branched bodies—and impose different demands on the controllers that operate them.

The BodyGen-based benchmark permits larger and more diverse morphologies than UNIMAL. Although both benchmarks initialize \textsc{Crawler} with an ant-like structure, the design space used here allows up to 29 limbs, compared with UNIMAL's limit of 10. Within the \textsc{Glider} and \textsc{Walker} families, progressively relaxed branching limits further expand the range of possible topologies while preserving each task objective. 
Together, the BodyGen-based environments test whether a co-design method can explore structurally diverse morphologies and learn compatible controllers under distinct task demands. By expressing topology edits and attribute updates as sequential actions evaluated through policy rollouts, the benchmark also provides a natural setting for RL-based morphology design and for assessing how effectively task feedback guides body--brain co-adaptation.

\paragraph{Search Methods and Computational Cost}
Methods using the UNIMAL design space address morphology-evaluation costs differently. DERL~\citep{gupta2021embodied} evolves individual morphologies and trains a controller for each candidate, making fitness evaluation expensive. LOKI~\citep{jeon2026convergent} reduces this burden by sharing controllers among clusters of similar morphologies and exploring new designs through dynamic local search. Both approaches nevertheless search over explicit morphology candidates, whose performance must be evaluated as exploration proceeds.

GeCode instead organizes exploration around a small set of anchors in the compact Morphogene space. Conditioned rollouts explore body--controller designs locally around each anchor, while performance-guided updates move the anchors toward promising latent regions. 
Because these rollouts also train the shared design and control policies and provide the returns used for anchor updates, GeCode requires neither candidate-specific controller training nor a separate evaluation loop for global latent-space exploration, thereby balancing effective co-design with negligible additional computational cost (see App.~\ref{sec:app_efficient} for a detailed efficiency analysis).

\subsection{Limitations and Future Work}
\label{app:limit}
In this section, we discuss the limitations of the current framework and outline directions for extending Morphogene-driven co-design.

\paragraph{(1) Scaling to highly diverse morphologies.}
GeCode shares a control policy across Morphogenes, allowing experience to transfer between related designs. Early in training, however, the explored morphologies may differ substantially in topology and dynamics. 
A shared controller with limited capacity may learn some bodies more readily than others, biasing their observed returns and, in turn, the performance-guided Morphogene updates.
GeCode partly mitigates premature concentration by incorporating the best locally generated morphology code into each anchor update, thereby preserving anchor-specific exploration. 
This does not fully remove biases arising from unequal controller adaptation. 
Future work could use Morphogenes to guide knowledge routing or modular policy selection~\citep{feng2026knowledge,xie2025divcontrol,xie2025kind}, enabling specialization across body structures while retaining the benefits of shared learning.

\noindent\textbf{(2) Shared Morphogene Spaces Across Tasks.}
The Morphogene space is constructed from morphological structure and geometry rather than task rewards, suggesting that it may capture body-level factors shared across tasks. In this study, we construct the space and train the policies separately for each environment. Although offline Morphogene-space construction requires no environment interaction and incurs modest computational cost, whether Morphogenes can transfer to new objectives or terrains remains an open question. Future work could investigate a shared Morphogene space across compatible design domains, together with task-adaptive conditioning that reuses body-level knowledge while adapting morphology generation and control.

\noindent\textbf{(3) Physical deployment.}
Like much of morphology--control co-design research, GeCode is evaluated in simulation using MuJoCo~\citep{todorov2012mujoco}. 
Extending such methods to physical robots remains challenging: exploring diverse body structures requires fabrication or reconfigurable hardware, while most available robot platforms have fixed morphologies, such as humanoid~\citep{cao2025humanoid, liu2026bio} and quadruped designs~\citep{han2024lifelike, long2024hybrid}.
Fabrication constraints, actuator limits, and differences between simulated and real contact dynamics further complicate deployment.
Our results demonstrate Morphogene-driven co-design in simulation rather than on hardware. 
Nevertheless, Morphogene's compact representation of coordinated body--brain designs could help identify promising structural features and control requirements, providing guidance for the design and fabrication of physical robots.

\end{document}